\documentclass[pdflatex,sn-mathphys-ay]{sn-jnl}% Math and Physical Sciences Numbered Reference Style

\usepackage{multirow}%
\usepackage{amsmath,amssymb,amsfonts}%
\usepackage{amsthm}%
\usepackage{mathrsfs}%
\usepackage[title]{appendix}%
\usepackage{xcolor}%
\usepackage{textcomp}%
\usepackage{manyfoot}%
\usepackage{booktabs}%
\usepackage{algorithmicx}%
\usepackage{algpseudocode}%
\usepackage{listings}%
\usepackage{lscape}

\usepackage{graphicx}
\usepackage{algorithm}
\usepackage{subcaption}
\usepackage{makecell}
\usepackage{tabularx}
\theoremstyle{thmstyleone}%
\newtheorem{theorem}{Theorem}%  meant for continuous numbers
\theoremstyle{thmstyletwo}%

\theoremstyle{thmstylethree}%
\newtheorem{definition}{Definition}%

\begin{document}
\setcounter{page}{0}

\title[Article Title]{Evolutionary Developmental Biology Can Serve as the Conceptual Foundation for a New Design Paradigm in Artificial Intelligence}


%\author{\fnm{Anonymous} \sur{Authors}}

\author*[1]{\fnm{Zeki Doruk} \sur{Erden}}\email{zeki.erden@epfl.ch}

\author[1]{\fnm{Boi} \sur{Faltings}}\email{boi.faltings@epfl.ch}

\affil*[1]{\orgdiv{Artificial Intelligence Laboratory}, \orgname{École Polytechnique Fédérale de Lausanne}, \orgaddress{\street{Route Cantonale}, \city{Lausanne}, \postcode{1015}, \state{Vaud}, \country{Switzerland}}}


\abstract{Artificial intelligence (AI), propelled by advancements in machine learning, has made significant strides in solving complex tasks. However, the current neural network-based paradigm, while effective, is heavily constrained by inherent limitations, primarily being a lack of structural organization and a progression of learning that displays undesirable  properties. As AI research progresses without a unifying framework, it either tries to patch weaknesses heuristically or draws loosely from biological mechanisms without strong theoretical foundations. Meanwhile, the recent paradigm shift in evolutionary understanding—driven primarily by evolutionary developmental biology (EDB)—has been largely overlooked in AI literature, despite a striking analogy between Modern Synthesis and contemporary machine learning, evident in their shared assumptions, approaches, and limitations upon careful analysis. Consequently, the principles of adaptation from EDB that reshaped our understanding of the evolutionary process can also form the foundation of a unifying conceptual framework for the next design philosophy in AI, going beyond mere inspiration and grounded firmly in biology's first principles. This article provides a detailed overview of the said analogy between Modern Synthesis and modern machine learning, and outlines the core principles of a new AI design paradigm grounded in insights from EDB. To exemplify our analysis, we also present two learning system designs grounded in specific developmental principles—regulatory connections, somatic variation \& selection, and weak linkage—that resolves multiple major limitations of contemporary machine learning in an organic manner, while also providing deeper insights into the role of these mechanisms in biological evolution.}




\maketitle

\section{Introduction}
\label{sec:introduction}

Over the past decades, artificial intelligence (AI) has advanced considerably, enabling the solution of complex, high-dimensional tasks—such as vision, behavior, and language processing—that were once thought impossible for \textit{classical (symbolic) AI} methods (e.g., classical computer vision, planning, or inference algorithms). However, the dominant paradigm of AI today—what we may call \textit{contemporary machine learning (ML)}—relies on slow, incremental statistical optimization of large, unstructured, overparameterized neural networks (\citep{bengio2017deep, du2018power}). While highly effective, this approach faces fundamental limitations that are becoming increasingly evident as its successes push against its boundaries (\citep{clune2019ai, sebastianrisiFutureArtificial, zador2019critique, marcus2018deep, lecun2022path}). There is a growing recognition of the need for a new design philosophy in AI. Yet, thus far, the field has lacked a clear, unifying conceptual framework to guide this search. Many proposed alternatives either attempt to patch the weaknesses of contemporary ML or integrate it with symbolic methods through various heuristics. When biology serves as inspiration, it is often in the form of loosely borrowed mechanisms—primarily from neuroscience—rather than an integrative understanding built upon first principles. As a result, the quest for AI’s future remains largely exploratory, lacking both strong theoretical foundations and a cohesive vision.

The past decades have also witnessed a revolution in evolutionary theory—one that extends the Modern Synthesis (MS), the dominant view of evolution throughout the 20\textsuperscript{th}-century, by integrating insights from diverse biological fields, addressing key explanatory gaps and fundamentally deepening our understanding of evolution. Among the most significant contributions are those of \textit{evolutionary developmental biology} (\textit{evo-devo} or \textit{EDB}), which have importantly shed light on rapid, accelerating evolutionary transitions and the structural properties of organisms that shape their adaptability and evolvability.

This conceptual shift in evolutionary theory has largely gone unnoticed by AI researchers, as the two fields have historically been seen as only tangentially related.\footnote{This statement specifically concerns the study of \textit{artificial} intelligence. In contrast, theories of "natural" intelligence have long considered deep connections between neural systems and evolutionary principles. Local variation and selection are fundamental to neural development (\citep{hiesinger2021self, nihDevelopingBrain, huttenlocher2013synaptogenesis, gu2013neurogenesis, gonccalves2016vivo, mowery2023adult}), and beyond local adaptation, evolutionary principles have been proposed to explain higher-order brain functions, including neural group selection (\citep{edelman-neuraldarwinism-1987, edelman1993neural}) and neuronal replicator dynamics (\citep{fernando2010neuronal, de2015neuronal, fedor2017cognitive}), among others (\citep{seung2003learning, adams1998hebb, changeux1973theory, loewenstein2010synaptic, calvin1987brain, calvin1998cerebral, fernando2012selectionist}).} However, a closer examination reveals otherwise: the philosophy of contemporary machine learning mirrors that of the Modern Synthesis in its assumptions, its conceptualization of adaptation, and its analytical scope. As a result, the explanatory gaps of MS correspond directly to the unresolved limitations of contemporary ML. This naturally positions this \textit{extended synthesis}\footnote{Throughout this paper, we use "Modern Synthesis" to refer to evolutionary theory as it stands without the conceptual contributions of developmental biology and other recent fields. We take no stance on whether these additions should be considered within the MS framework or as a distinct theoretical shift, but we adopt this distinction for conceptual clarity.}, and particularly evolutionary developmental biology, as a strong candidate for a unifying conceptual foundation for AI—one that moves beyond superficial biological analogies and instead grounds itself in fundamental conceptual principles of adaptation and biological organization instead. Conversely, this connection also highlights AI’s potential as a primary computational modeling framework for EDB, offering a platform to test and refine theories linking adaptation and developmental principles. Just as AI has contributed to neuroscience (\citep{slack2023dalle, macpherson2021natural, maheswaranathan2023interpreting}) and MS-based evolutionary modeling (\citep{levin1995evolution, birchenhall1997genetic}), it could provide new insights into the adaptive processes central to EDB. Recognizing this organic relationship—where both fields study adaptation through shared underlying principles—suggests the need for a tighter integration of AI and evolutionary theory, with the potential to transform both disciplines.

This paper explores the connection outlined above in detail. In Section \ref{sec:evol_ai}, we provide a conceptual overview of contemporary AI technologies, accessible to both newcomers and those already acquainted with the field, analyzing the strengths behind their recent success as well as their fundamental limitations. In the same section, we also provide an overview of Modern Synthesis and highlight the conceptual analogy of contemporary AI and this understanding of evolution. Section \ref{sec:evolution} then introduces the theoretical framework that EDB offers for a new design paradigm for AI. To substantiate our arguments, we present two artificial learning system designs built on fundamental developmental principles we emphasize throughout the earlier parts: regulatory connections (Sec. \ref{sec:dirad}) and adaptation via somatic variation \& selection with weak linkage (Sec. \ref{sec:Modelleyen}). We demonstrate how these principles address inherent limitations of current learning systems with theoretical guarantees while also offering potential insights into their biological counterparts. Our argumentation presents a unified conceptual framework that can serve as a foundation for the future of both AI and EDB, with a closer connection between the two disciplines, with the demonstrations providing concrete examples of these principles in action, illustrating their transformative potential for both fields.

\section{The Parallels Between Evolutionary Theory and the State of AI}
\label{sec:evol_ai}

\subsection{Contemporary machine learning}
\label{sec:contemporary_ai}

Modern AI over the past 10-15 years has demonstrated remarkable success in tasks that were once beyond the reach of traditional computer science algorithms, including methods such as planning and classical computer vision, which were historically regarded as core components of "AI methods" and formed the dominant paradigm prior to the rise of modern techniques. We will present a summary of the design paradigm that underpins contemporary AI, with a specific focus on machine learning, the cornerstone of most prevailing approaches. We will explore the capabilities of these systems, the factors contributing to their success, and their inherent limitations.

Modern machine learning is primarily based on the concept of function approximation, where the system's desired output is treated as a function to be approximated, or "learned." The dominant method in contemporary AI for achieving this is artificial \textit{neural networks (NNs)}, particularly a subclass known as \textit{multilayer perceptrons (MLPs)} (\citep{bengio2017deep}). Given the pervasive adoption of MLPs in machine learning, we adhere to the common practice in AI literature of using these terms interchangeably throughout this discussion.

\begin{figure}
    \centering
    \includegraphics[width=0.7\linewidth]{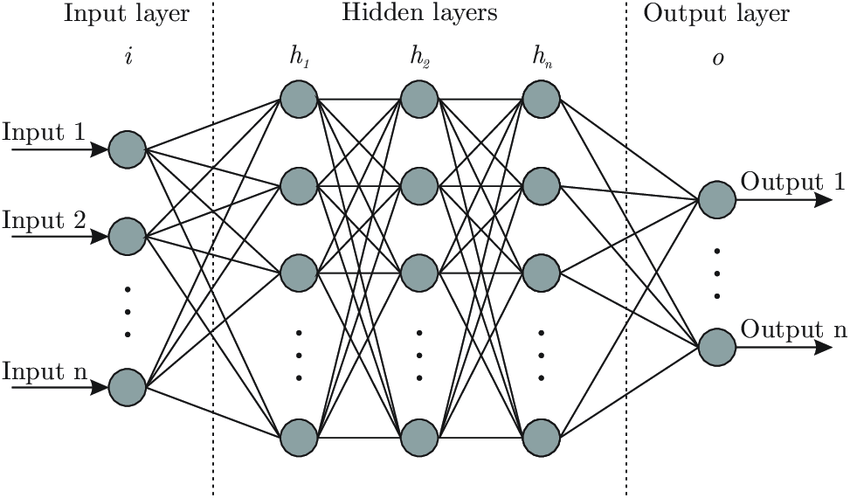}
    \caption{(From \citep{bre2018prediction}) A visualization of a \textit{multilayer perceptron}, the overwhelmingly dominant type of \textit{artificial neural network} in contemporary machine learning. This network processes data through an input layer, applies transformations via weighted computations across hidden layers, and ultimately produces target outputs. Learning occurs as the response biases of the network's nodes (\textit{neurons} or \textit{perceptrons}) and the weights of the connecting edges are iteratively adjusted to minimize an externally defined error.}
    \label{fig:ann_visualization}
\end{figure}

An MLP consists of a large number of simple, homogeneous subcomponents, referred to as \textit{artificial neurons} (with the questionable supposition of a reasonable analogy with biological neurons (\citep{katyal2021connectionismcomplexitylivingsystems, iyer2022avoiding})) or \textit{perceptrons}, and the connections between them (see Figure \ref{fig:ann_visualization}). Mathematically, the response $y$ of a neuron is defined as:

\begin{equation}
    \label{eq:nn_y}
    y = f(z) = f(\sum_{i} w_i * x_i + b)
\end{equation}

where $f(\cdot)$ is a nonlinear function, $x_i$ is the input to the perceptron (either externally or from other neurons), $w_i$ is the weight connecting $x_i$ to the neuron, and $b$ is a bias.  $w_i$ and $b$ are learnable parameters, initially assigned small random values, and are gradually adjusted to minimize the approximation error between the predicted and desired output functions. This adjustment is carried out through a process known as gradient descent, with the following update rule applied at each step:

\begin{equation}
    \label{eq:nn_grad_update}
    \alpha_{new} = \alpha - \gamma \frac{dC}{d\alpha}
\end{equation}

$\alpha$ being the parameter ($w_i$ or $b$), $\gamma$ being a preset parameter known as the learning rate, and $C$ representing the cost function of the network, which decreases as the NN's output gets closer to the externally defined target functions (which may be e.g., one-hot vectors encoding different objects for a network trained for object detection). The gradient of the cost with respect to each parameter, $\frac{dC}{d\alpha}$, is computed using a straightforward algorithm (backpropagation (\citep{bengio2017deep})) that applies the chain rule across the entire network, the details of which are beyond the scope of this paper. Intuitively, Eq. \ref{eq:nn_grad_update} means that the learnable parameters gradually move along the slope of a fixed "error function landscape" (analogous to a simple form of a fitness landscape in evolutionary theory (\citep{fragata2019evolution})), descending towards \textit{a local minimum.}  Importantly, the cost function is typically defined as $C = \frac{1}{m} \sum_m C_m$, where $C_m$ represents the cost (mismatch between network output and target responses) associated with each individual example $m$ in the task the network is trained on (e.g., a particular image of a particular object). Thus, $C$ reflects the \textit{average} mismatch across the entire dataset. Consequently, the gradient $\frac{dC}{d\alpha}$ reflects the \textit{net influence} of the whole dataset on the learned parameters, meaning that learning proceeds to a point that reflects an adaptation to the dataset \textit{on average}. This can result in \textit{statistical trade-offs}, where conflicting gradients lead to learned values that represent mid-points that minimize this average error. Universal approximation theorems (\citep{cybenko1989approximation}) assert that neural networks can approximate any function given a sufficient number of parameters, which in practice results in extreme overparameterization (\citep{du2018power})—a redundancy that is necessary because gradient descent requires ample degrees of freedom to escape local minima representing suboptimal stable trade-offs among the competing pressures of multiple examples, and to instead converge on solutions that better fit the majority of the training data.

\begin{figure}
    \centering
    \includegraphics[width=0.7\linewidth]{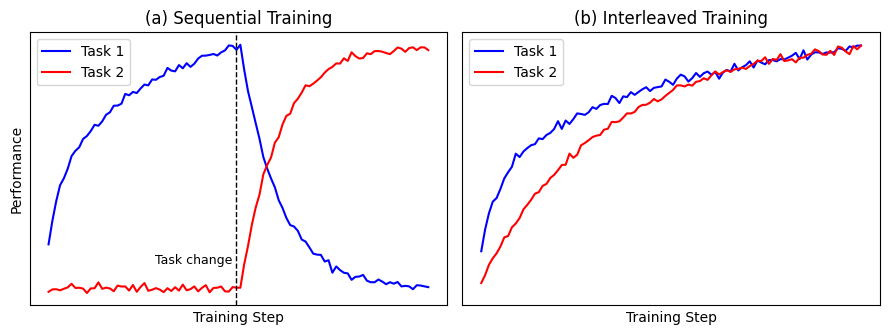}
    \caption{Illustration of the destructive adaptation problem with \textit{continual learning}, recreated from \citep{vandeven2024continuallearningcatastrophicforgetting}. \textit{(a)} Sequential training in multiple tasks or environments results the neural network to destroy knowledge of the previous ones, visible as a sharp decrease in performance. \textit{(b)} This is not the case if the network learns multiple tasks in parallel or "replayed" the data pertaining to the previous task when learning a new one, which shows that destruction of past knowledge is not due to any inherent capacity limit or an incommensurability of the tasks/environments.}
    \label{fig:destructive_adaptation}
\end{figure}

Methods from the previous design paradigm of AI, often referred to as “classical” or “symbolic AI” (which includes techniques such as planning, reasoning, and traditional computer vision), often struggled to address complex, high-dimensional tasks such as language processing, analyzing intricate images, or managing behavior in highly unpredictable environments. In recent decades, large-scale neural networks have become feasible to train, primarily due to advancements in computing hardware (\citep{khan2021advancements, baji2017gpu}) and the widespread availability of domain-specific data (\citep{duan2019artificial}). As a result, many tasks that were previously unsolvable using classical methods are now feasible through neural networks (\citep{khan2021machine, chai2021deep, zhao2023survey, min2023recent, li2017deep}), provided sufficient training data is available. However, as these modern methods steadily conquer challenges once thought insurmountable, growing attention is being drawn to their fundamental, inherent limitations and their ability to integrate into broader systems (\citep{clune2019ai,zador2019critique,marcus2018deep,lecun2022path}). We think that these limitations can be categorized in two main categories:

\begin{enumerate}
    \item \textbf{Obliteration of the existing knowledge when learning new information:} Modern ML methods are incapable of effectively integrating new knowledge into existing knowledge without destroying old information, which is crucial for the capability of open-ended \textit{continual learning} (\citep{vandeven2024continuallearningcatastrophicforgetting, hadsell2020embracing}). This limitation not only forces AI systems to undergo extensive retraining and reinteraction with the environment where such interaction is not feasible, but also eliminates their prospects to expand their knowledge and capabilities in an adaptive and recursive manner. How old knowledge is destroyed when neural networks learn new tasks will be crucial for our argument: NNs approximate a function that represents the \textit{average} effect of a large dataset, which is only aligned with the desired essence of the task when all relevant data is considered \textit{together}. In large, overparameterized networks, this enables impressive performance, provided that \textit{the data is available from the outset}. However, when a new task is introduced, the network does \textit{not} create a combined objective function; instead, it optimizes \textit{exclusively} the new target function, with no pressure to retain performance on previous tasks. This leads to the destruction of previously learned knowledge (represented by existing weight patterns) when faced with new tasks, manifesting as a sharp decline in performance on the old task (\citep{vandeven2024continuallearningcatastrophicforgetting}) as illustrated on Figure \ref{fig:destructive_adaptation}. Several \textit{ad-hoc} fixes exist for this problem of \textit{destructive adaptation}\footnote{This phenomenon is often referred to as "catastrophic forgetting" in AI literature. However, this term is misleading as it implies an unintentional loss of unused knowledge. In contrast, the actual process involves the active destruction of existing knowledge when learning a new task. To avoid this misinterpretation, we refer to this phenomenon as "destructive adaptation" instead}, (see e.g. \citep{rolnick2019experience, buzzega2021rethinking, buzzega2020dark, galashov2023continually, masse2018alleviating, jacobson2022task, kirkpatrick2017overcoming, wang2022continual, iyer2022avoiding, rusu2016progressive, erden2024directed, lee2020neural, erden2024directed, lee2020neural, iyer2022avoiding}), but all of them come with important constraints that clash with the needs of real-world learning and act as mitigations of the issue, rather than proper solutions.\footnote{For example, some methods rely on storing and replaying past examples, which is straightforward but becomes impractical over the long lifetimes of intelligent agents (\citep{rolnick2019experience, buzzega2021rethinking, buzzega2020dark, galashov2023continually}). Other approaches depend on clearly defined boundaries between different "tasks" or "environments" (\citep{masse2018alleviating, jacobson2022task, kirkpatrick2017overcoming, wang2022continual, rusu2016progressive, erden2024directed, lee2020neural, iyer2022avoiding}), often visible to the agent, which is a highly unrealistic in the real world.}

    \item \textbf{Incomprehensibility and non-engineerability:} The second critical limitation of modern ML methods is the incomprehensibility of AI systems' internal structures, which hinders their widespread adoption and poses significant risks in senstitive scenarios (\citep{marcus2018deep}). NNs rely on overparameterization, deriving representations from continuous parameters fine-tuned to a few significant digits and exhibiting nonlinear responses from hundreds or thousands of inputs per neuron. This not only results in a learned structure that is completely incomprehensible, hence making these systems inherently nontrustworthy and impossible to analyse in detail; but also makes engineerability of these systems (such as validation, incorporation of designer knowledge, or utilization of learned representations in other, non-learning processes) impossible. This is a very important limitation, as "classical AI," which cannot be easily integrated with contemporary ML techniques due to this lack of engineerability, is backed by decades of research and provides robust solutions for tasks that cannot be efficiently addressed by data alone. Currently, no mainstream method can reliably incorporate a comprehensible, decomposable, and engineerable structural organization of ML systems across all levels of organization (\citep{su2024focuslearn, goyal2020object, pateria2021hierarchical, andreas2017modular, devin2017learning, sahni2017learning, goyal2019reinforcement, yang2020multi}),\footnote{To summarize, many methods that attempt to incorporate structure to NNs impose rigid, predefined substructures from above that lack dynamism, rather than generating structured organization themselves (\citep{su2024focuslearn, goyal2020object, pateria2021hierarchical}). Meanwhile, approaches that attempt to circumvent this constraint to some degree often rely on unstable foundations that restrict their expressive capacity due to inherent limitations in contemporary ML systems, such as dynamic task-specific substructures in continual learning being fundamentally constrained by the task-dependency issues of NNs (pt. (1) above) (\citep{andreas2017modular, devin2017learning, sahni2017learning, goyal2019reinforcement, yang2020multi}). Moreover, no existing approach achieves structural organization across all levels of organization in a fractal-like manner, limiting benefits to a layer atop the underlying neural networks while accepting a monolithic representation below this level.} and the research on explaining internal operations mainly focuses on post-hoc analyses of responses and parameter statistics rather than addressing the systems' intrinsic complexity (\citep{xu2019explainable}). This situation stands in stark contrast to nearly all human-engineered systems. While it may be unrealistic to expect a system trained on vast amounts of data to encapsulate all of that knowledge with a small number of parameters or to be immediately comprehensible at every level, it is reasonable to expect a degree of comprehensibility at higher levels of abstraction or within each organizational layer. This would mirror biological systems, which, despite their immense complexity, remain fundamentally understandable when analyzed at the appropriate level of abstraction. This lack of clarity hampers our understanding of how these systems function and inhibits decomposability and low-level modifiability, making AI inherently uncontrollable and unreliable beyond its statistical guarantees.
\end{enumerate}

These limitations of contemporary AI are well acknowledged—sometimes in isolation, sometimes together as a whole—but rarely addressed in unity with a solid conceptual foundation. Efforts to address these issues often take the form of patchy mechanisms layered on top of existing systems or calls for redesign without explicitly specifying the foundational principles that should guide such change. This fragmented approach prevents a coherent resolution, leaving fundamental challenges unresolved.

\subsection{Modern Synthesis and its limits}
\label{sec:modern_synthesis}

Darwinian theories of selection \citep{darwin1964origin}, combined with Mendelian genetics and modern understanding of genetic variation has formed the understanding of evolutionary biology that has dominated 20\textsuperscript{th} century, termed the Modern Synthesis (MS) \citep{huxley1942evolution, gilbert2000developmental, mayer2013evolution}. MS views evolution as driven by inheritable \textit{genetic variation}, arising from mutations and recombination, enabling populations to adapt to environmental changes. Selection, acting on \textit{phenotypic variation}, favors traits that enhance survival and reproduction. MS treats evolution as a process at the population level, abstracting away the complex internal processes of organisms and development, which aren't central to its explanation \citep{walsh2015organisms, gilbert2000developmental, marc2005plausibility, carroll2005endless, laland2015extended}. Biological evolution is understood as a \textit{statistical} property of populations rather than individual organisms; under the strict Modern Synthesis framework, it is expected to proceed through slow, gradual shifts in a population’s genetic makeup \citep{johnston2019population, carroll2005endless}, with also no inherent reason to anticipate an accelerating rate of phenotypic change over time.

Modern Synthesis has been immensely successful in advancing our understanding of evolution, with many practical implications \citep{gould2009antibiotic, dolgova2018medicine, thrall2011evolution, olivieri2016evolution}. But as powerful as it is, and as much as it provides an accurate conceptual understanding of the high level dynamics of the evolutionary process; in its pure form it leaves critical gaps in its ability to explain aspects of evolution, whose importance are now understood to be much more important than initially thought: Firstly, the MS does not incorporate (neither as explanandums nor as explanans) the generation of biological structures \citep{walsh2015organisms, gilbert2000developmental, marc2005plausibility, carroll2005endless} that exhibit widespread features like \textit{repetition/reuse} \citep{preston2011reduce, anderson2010neural, barthelemy1991levels}, \textit{correlation} \citep{paaby2013many, watson2014evolution, price1992evolution}, \textit{modularity} \citep{lorenz2011emergence, clarke1992modularity, kadelka2023modularity, clune2013evolutionary, wagner2007road}, and \textit{hierarchical composition} \citep{mengistu2016evolutionary, ingber2003tensegrity, uversky2021networks, grene1987hierarchies, pan2014exploring} in phenotypes. By viewing evolution as a gradual optimization of allele frequencies, MS excludes the internal complexity of organisms from its inquiry, deeming it supposedly irrelevant for understanding evolutionary processes and abstracting it away within genetic variation. This perspective treats organisms more as mere vessels for genes than as intricate entities with their own developmental processes \citep{walsh2015organisms}. Secondly, the qualitative pattern of evolutionary progression predicted by Modern Synthesis differs from what is observed. Evolution is not merely a steady, linear process; it also includes periods of \textit{rapid transformation} \citep{gould1977punctuated, marc2005plausibility} and more notably, an \textit{accelerating} increase in the speed of morphological change and phenotypic complexity throughout evolutionary history \citep{smith1997major, koonin2007biological, russell1983exponential, sharov2006genome, heylighen2000evolutionary, kurzweil2006singularity}. To exemplify, the timeline of life's complexity—from the emergence of life ($3.5$ bya) to eukaryotes ($2$ bya), multicellular organisms ($1$ bya), and the Homo genus ($3$ mya) (see also Table 1 in \citep{gerhart2007theory}) reveals that while these exemplary events occur over a roughly linear timescale, phenotypic complexity has increased exponentially \citep{vosseberg2024emerging, grosberg2007evolution, ispolatov2012division}.

The underlying issue is that the Modern Synthesis effectively addresses two of the foundational pillars of evolutionary theory —selection and inheritance— but provides a limited account of the third pillar, variation \citep{marc2005plausibility}. While it clarifies the origins of \textit{genetic variation}, it treats the transition to phenotypes as a black box, failing to fully explain \textit{phenotypic variation}. This leaves much of the "how" of evolution unaddressed \citep{marc2005plausibility, laland2015extended}. To illustrate, an analogy can be drawn between Modern Synthesis and current mainstream evolutionary computation methods like basic genetic algorithms, which some readers may be more familiar with compared to evolutionary biology. These algorithms excel at optimizing predefined structures, such as selecting the best variable combinations or tuning design parameters \citep{bayley2008design, chiroma2017neural}, but in their basic form they cannot spontaneously generate large-scale functional structures or implement rapid, large adaptive changes. As a support to the argument that will follow in the coming sections, the most notable exception to this within the evolutionary algorithms literature is a class of methods called \textit{indirect encodings} \citep{stanley2003taxonomy, meli2021study, gauci2010indirect} which fall under the category of methods that incorporate mechanisms inspired by or based on developmental or embryogenic processes. (See also the related footnote in Section \ref{sec:evolution}.)

In recent decades, advancements in biology have expanded Modern Synthesis, addressing gaps in this framework \citep{laland2015extended}. Among these, evolutionary developmental biology (EDB) has been particularly impactful \citep{carroll2005endless, marc2005plausibility, west2003developmental, gerhart2007theory}, focusing on how developmental mechanisms drive evolutionary patterns. Before we dive into these insights, however, we must motivate this by outlining the analogy between Modern Synthesis and the philosophy \& design of the current approaches in artificial learning systems.

\subsection{The analogy of Modern Synthesis and contemporary machine learning}

The Modern Synthesis of evolution and contemporary machine learning exhibit striking parallels in terms of their assumptions, their understandings of adaptive tasks \& the processes of adaptation, supported also with a compelling analogy that can be drawn between the \textit{explanatory assumptions and unresolved gaps} of the former and the \textit{capability constraints and functional shortcomings} of the latter.\footnote{The analogy we present should not be mistaken for a simplistic, one-to-one mapping between the fundamental conceptual elements of the evolutionary theory as represented by Modern Synthesis—such as populations, organisms, genotypes, and so forth—and their direct counterparts in machine learning. While some ML methods that clearly fall within the current paradigm (e.g., standard genetic algorithms; see above) do permit such straightforward correspondences, the discussion we develop here—rooted, as will become clear, in the \textit{philosophy, conceptual framework,} and \textit{domain of interest} of both fields—extends well beyond such cases; crucially also encompassing non-population-based machine learning approaches, such as the dominant \textit{gradient descent} paradigm. It is also worth emphasizing that this argument is not limited to machine learning alone but applies more broadly to many standard iterative optimization techniques. Nonetheless, we confine our focus to ML, both due to its central relevance in our context and because these issues are especially acute in AI, where adaptivity and learning from limited data are essential demands of the domain.} These parallels, which attentive readers may have already identified in the earlier discussion, can be explicitly outlined under two categories (summarized in Table \ref{tab:ms_ml_analogy}):

\begin{itemize}
    \item \textbf{\textit{The progression pattern of the adaptive process:}} In the Modern Synthesis perspective, evolution is viewed as a slow, gradual process, characterized by incremental changes in the genetic composition of populations \citep{johnston2019population, carroll2005endless}. Major changes are typically considered detrimental, and there is no expectation that past adaptations will enhance future evolvability and drive a recursively accelerating pace of change. This view conceptualizes the adaptive process as a gradual search over a rugged fitness landscape, progressing through random sampling of nearby regions in close vicinity. This stands in contrast to the qualitative pattern of evolutionary progression, which involves periods of rapid transformation, and more notably, an acceleration in the rate of morphological change and phenotypic complexity throughout evolutionary history. Analogously, in contemporary ML, learning is viewed as a gradual process of minimizing a target function by following the gradient towards lower values in the immediate vicinity of the current solution, occurring slowly over many iterative steps across the training data. Existing learned knowledge offers little benefit for future learning unless tasks are highly similar (in which case, it can provide only a limited advantage as a good initialization point for new tasks), and is often destroyed when new knowledge is to be incorporated. Open-ended learning with the capacity for recursive improvement is not observed, nor is there a reasonable expectation for it to emerge. Furthermore, in the Modern Synthesis framework, organisms are prone to getting stuck in local adaptive peaks in this rugged landscape unless large environmental changes force them out of these peaks. This phenomenon is mirrored in machine learning, where considerable overparameterization is required to avoid settling in a noticeably inferior local minimum.

    \item \textbf{\textit{Exclusion of internal structure from domain of interest:}} MS treats evolution as a process at the population level, abstracting away the complex internal processes of organisms and development, which aren't central to its explanation \citep{walsh2015organisms, gilbert2000developmental, marc2005plausibility, carroll2005endless, laland2015extended}. It views its subjects as largely unstructured entities whose relevant properties are captured within the allele frequencies, and emphasizing a numerically-defined "fitness" as the main signifier of the process. Consequently, MS does not incorporate the \textit{generation of biological structures} that exhibit properties of \textit{repetition, correlation, modularity, or hierarchical composition} in phenotypes. Analogously, contemporary ML relies on unstructured, overparameterized networks as the foundation for executing learning tasks. Relevant information is encoded in thousands-to-billions of continuous weight values, whose semantic meaning—and by extension, the semantics of the task or environment the system is learning—are considered irrelevant for the learrning capability. Any potential structural organization that might exist within the target task (such as core repeatable functionalities, independent or isolated components, or the multi-level decomposability of goals) is assumed to be implicitly embedded within this monolithic network, rather than being considered part of the design or analysis domain. As a result, the learned models become unstructured (failing to embody the properties of a \textit{modular, hierarchical, repetitive, and correlated structural organization}, the very same properties that the Modern Synthesis, on its own, does not explain the generation of) and, by extension, incomprehensible and non-engineerable.
\end{itemize}

The striking resemblance between the previously dominant view of evolution and contemporary ML, particularly with respect to their limitations, raises a compelling question: Can the principles behind the conceptual shift in evolutionary theory over the past few decades, specifically those of evolutionary developmental biology, serve as a foundation for overcoming the major inherent limitations of the current AI design paradigm? More precisely, can these principles provide a strong, unifying conceptual framework—grounded in first principles (rather than an often-superficial inspiration from specific biological mechanisms)—that can guide the field, which currently approaches solutions to its problems in a disconnected and largely heuristic manner, over the coming decades? We believe the answer to these questions is affirmative. In the following section, we provide a brief overview of the principles and insights from EDB that address the limitations of the Modern Synthesis, which are analogous to those of contemporary ML. We will then discuss how these insights can be translated into design principles for a new paradigm in AI system development.\footnote{The argument we will outline should not be confused with the view that AI systems must evolve a "genome" and express that genome as the final system, akin to a "phenotype" in the genotype-phenotype transition. While some approaches already explore this (e.g., \citep{wilson2022evolving, najarro2023towards, zhang2024evolved}), and an argument in their favor can certainly be made—particularly given their potential long-term ramifications—evolving systems in this manner remains costly and has yet to demonstrate a clear advantage in addressing the core limitations of contemporary ML systems that we discussed (and many of these approaches do not explicitly position themselves as solutions to these limitations in the first place) over a well-designed learning approach that directly modifies the learned model while inherently incorporating the same principles into its design. Most of these evolutionary-developmental approaches in AI are better characterized as "inspired by specific mechanisms in biology" (albeit toward the more substantiated side) rather than what we refer to as "based on a unifying conceptual framework grounded in foundational first principles." Although evolutionary developmental biology is closely tied to the concept of genotype-phenotype mapping, the general first principles of adaptation emerging from the field \textit{do not} require a system to be separated into a genotype and its expression as a phenotype, as we will demonstrate in the discussion below.}

% Removed from above: "And further, can AI systems be used to inform or model conceptual studies in EDB (and the broader extended evolutionary theory (\citep{laland2015extended, mcghee2023evolutionary})), in the same way they have contributed to neuroscience research until now (\citep{slack2023dalle, macpherson2021natural, maheswaranathan2023interpreting})?"

\begin{landscape}
    \begin{table}
        \centering
        \caption{The analogy between the Modern Synthesis and contemporary Machine Learning, in terms of their conceptual foundations and the limitations arising from these foundations. Assumptions of the Modern Synthesis include, in parentheses, the corresponding "core limitations" as categorized in \citep{laland2015extended}, Table 1.}

        \begin{tabularx}{\linewidth}{X|X|X|X}
            \multicolumn{2}{c|}{\textbf{Modern Synthesis}} & \multicolumn{2}{c}{\textbf{Contemporary Machine Learning}} \\
            \textbf{Assumption} & \textbf{Resultant explanatory gap} & \textbf{Conceptual approach} & \textbf{Resultant technical limitation} \\
            \hline
            Internal structure of the organism is irrelevant for understanding evolutionary progression. Relevant properties for adaptation can be captured in gene frequencies, with the translation to phenotype considered unimportant. \textit{(Gene-centered perspective)} & The generation of the phenotype, including widely observed structural properties of organisms (e.g., modularity, hierarchy, repetition, correlated changes), remains unexplained. & Learning can be performed with unstructured, monolithic, overparameterized networks. Relevant information is encoded in network weights, with the semantic meaning of these weights irrelevant for design purposes. & Learned models, embedded in unstructured networks, are non-decomposable, incomprehensible, and non-engineerable.  \\
            \hline
            Evolution is a process of gradual search over a rugged fitness landscape that proceeds with random sampling of the near vicinity and traversal in the direction of higher fitness. \textit{(The pre-eminence of natural selection, random genetic variation, gradualism, macro-evolution)} & (1) Evolution is expected to occur slowly, as mutations with large effects are likely to disrupt existing structures.

            (2) There is no fundamental mechanism by which past adaptations can benefit future evolvability.
            
            (3) Periods of rapid change, as well as a recursively accelerating increase in phenotypic complexity and capabilities, remain unexplained.
            
            (4) Organisms are prone to being stuck in local adaptive peaks in the absence of changes that drive them out of these peaks. & Learning is a process of gradual minimization of a target function based on the gradient towards lower values in the immediate vicinity of the current solution. & (1) Learning occurs slowly, across many iterative steps over the same training data. Learning new knowledge destroys existing knowledge.
            
            (2) Existing learned representations offer no benefit for future learning, other than possibly serving as a good initialization point if previous and new tasks align well, and are often obliterated when learning new representations.
            
            (3) Open-ended learning with recursively improving capabilities is not observed.
            
            (4) A solution that does not fall into an inferior local minimum can only be achieved with significant overparameterization.
            \\
            \end{tabularx}
        \label{tab:ms_ml_analogy}
    \end{table}
\end{landscape}

\section{EDB as the conceptual foundation of a new design philosophy}
\label{sec:evolution}

\subsection{The conceptual basis from evolutionary developmental biology}

EDB dispels the notion that genotype-to-phenotype transformation is irrelevant to understanding evolution, demonstrating that universally shared developmental principles not only drive organismal complexity and adaptability but also illuminate core aspects of evolution itself. The field has made numerous major contributions to evolutionary theory, many of which hold clear potential for informing the design of intelligent systems and other engineering disciplines. However, certain insights from EDB are of particular relevance to us here, given their power to address key limitations of the Modern Synthesis and hence, offer direct implications for contemporary AI. These are detailed in the following three subsections, with a summary and synthesis in Section \ref{sec:ai_future}.

\subsubsection{\textbf{The structure of gene regulation}}
\label{sec:edb_structure_genereg}

A central insight of EDB is that the generation of phenotype is not solely driven by the coding genes and proteins an organism possesses, but is primarily influenced by changes in the regulatory genes that control gene expression. These regulatory genes themselves evolve through genetic switches that govern and condition gene expression \citep{Alberts2002, goncalves2012extensive, vaishnav2022evolution, wray2007evolutionary, roff2011alternative}, acting analogous to an "if condition" of biology. Examples include changes in regulatory elements that drive morphological novelties \citep{burgess2016sonic, tickle2017sonic, rebeiz2017enhancer} and the reconfiguration of neural gene expression networks that enhanced brain function in humans \citep{wang2016divergence, patoori2022young, suresh2023comparative}. Small changes in regulation, when unfolded across time during development, can lead to dramatic alterations in the adult organism’s phenotype, as strikingly exemplified by the induction of ectopic eyes in \textit{Drosophila} \citep{halder1995induction} as well as human polydactyly \citep{lettice2005preaxial}; each of which can be understood as the result of activating substructure-generating processes in locations where those processes are not typically expressed. Moreover, these regulatory genes can control the expression of other regulatory genes, enabling the hierarchical regulation of entire pathways of gene expression, where master regulators can initiate downstream cascades \citep{hansen2015effects}.

There are different reasons for the evolution of a regulatory mechanism for a gene, however one of these reasons will be particularly important for our discussion. That is their contribution to developmental plasticity throughout allowing conditioning and further downstream evolution of \textit{alternative phenotypes} without negatively interfering with each other. This allows for evolutionary innovations in non-detrimental ways, thereby allow organisms to avoid being trapped in local adaptive peaks \citep{west2003developmental, wagner2011origins, moczek2011role, poelwijk2007empirical}. Specifically, in environments where conflicting adaptive pressures arise from multiple possible conditions the organism may encounter, regulatory conditioning enables the emergence of alternative phenotypes. Rather than being locked into a single phenotype that merely represents a compromise between these opposing pressures, the organism can maintain a repertoire of phenotypes, each of which can be selectively expressed in response to specific environmental cues. These alternatives can be conditioned on various different environmental factors, such as status (e.g. aggressive vs. nonaggressive mating strategies), multiple-niches (e.g. camufoulaging wing colors in some butterflies), or stress (e.g. smaller vs. larger body sizes based on food scarcity) – numerous examples of this phenomenon, and an in-depth discussion of the importance of them in evolution, are listed in \citep{west2003developmental}. From an optimization perspective, what occurs in such cases is effectively a modification of the original parameter space through the introduction of a new parameter via regulatory mechanisms. For a visual illustration, the reader may refer to Figure \ref{fig:dirad_landscape} in Section \ref{sec:dirad}, which demonstrates how regulation alters the adaptive landscape of a single-parameter system within the developmental machine learning framework introduced in that section. While the effects of biological regulation on evolution are not as easily visualized, the underlying mechanism is analogous.

Importantly, the fundamental structural properties that biological organisms consistently display (Table \ref{tab:ms_ml_analogy}) stem primarily from the architecture of the regulatory networks (\citep{carroll2005endless, alcala2021modularity, verd2019modularity, wagner2011pleiotropic}), such as reuse of regulatory mechanisms across various processes (repetition/correlation), distinct gene sets regulating different parts of the organism (modularity), or higher-level regulatory genes controlling the expression of downstream genes (hierarchy).\footnote{While this text focuses on the generation of structural properties in organisms primarily through the lens of regulatory network architecture, the role of other factors—such as biophysical properties \citep{forgacs2005biological}—must also be acknowledged within the broader framework of evolutionary developmental biology.} These phenotypic structural properties of organisms reflect the nature and organization of gene expression within the genome. They also significantly influence how evolution alters organisms, facilitating the development of new functional structures while reducing the likelihood of detrimental changes: Modularity and hierarchical organization facilitate both correlated changes (where modifications to one part of the organism trigger rapid, synchronous changes in others) and isolated changes (allowing alterations in one component without affecting others) (\citep{carroll2005endless, verd2019modularity, clune2013evolutionary, mengistu2016evolutionary, seki2012evolutionary, eliason2023early}). Repetition of substructures, on the other hand, allows for the rapid generation of new structures from moderately developed origins (\citep{wagner2011pleiotropic, carroll1995homeotic}). Furthermore, they facilitate the preservation of ancestral traits (\citep{szilagyi2020phenotypes, carroll2005endless}), by rapid deactivation switch points responsible for expressing these traits without completely dismantling the underlying genetic machinery, potentially allowing for rapid re-emergence of a past, functional trait.

\begin{figure}
    \centering
    \includegraphics[width=0.75\linewidth]{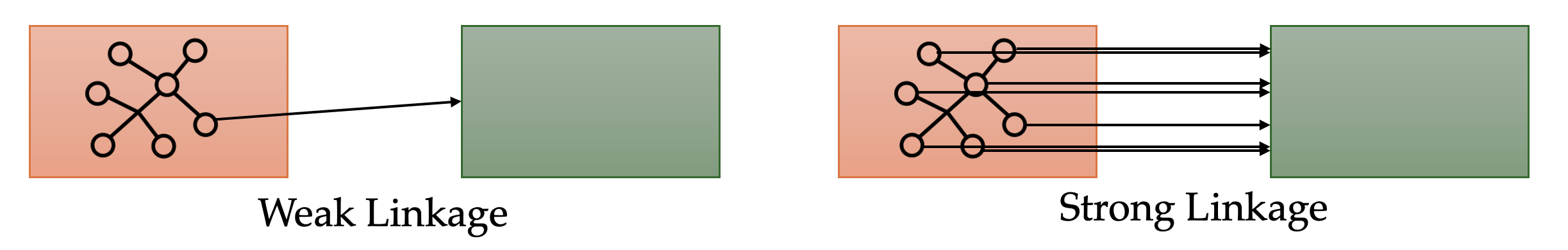}
    \caption{An illustration of \textit{weak linkage} (characterized by simple and sparse regulatory signals across components) and its alternative, which we may refer to as "strong linkage," marked by high interdependency, complex signal structures, and strong sensitivity to informative connections.}
    \label{fig:weak_linkage}
\end{figure}

An important concept in this context is \textit{weak linkage}, which refers to the simplicity of regulatory signals (inputs/conditions) that govern distinct processes—standing in contrast to complex, highly sensitive instructions that require precise integration and tightly connect multiple internal components across subsystems (see \citep{gerhart2007theory, marc2005plausibility}; illustrated in Figure \ref{fig:weak_linkage}). This simplicity greatly facilitates the evolution of regulatory changes and, more crucially, enables the combinability and reconfigurability to generate more complex functions. A prime example of this is the evolution of the vertebrate eye (\citep{lamb2007evolution, lamb2008origin}), where pre-existing components like photoreceptor cells (originally used for simple light detection, such as in phototaxis), the lens (derived from the dermis), and visual neural circuitry (derived from ancestral sensory neural circuits) were repurposed. This type of evolution requires fewer changes than constructing entirely new structures from scratch.

The contrast between these "design principles" and those in contemporary ML is clear. First and foremost, the fixed computational structures of machine learning systems preclude the presence of anything functionally analogous to the regulatory connections found in biological organisms: In neural networks, there is no mechanism for forming regulatory connections or switch points that govern the "expression" of learned (adapted) components, being the individual edge weights,\footnote{The fact that "edges regulate the firing of artificial neurons" should not be misleading in this regard. From an evolutionary perspective, artificial neurons themselves are \textit{not} adapted structures or "traits" (except for their thresholds, which represent a very limited amount of information). Majority of learning in a neural network takes place in the edge weights, hence any information patterns that actually represent something and are useful for an agent are represented within the edges of the network. Therefore, meaningful regulatory adaptation can only occur at the regulation of these edges or further upstream.} hence lacking the very basic mechanism to derive the benefits that the organization of the genome as a regulatory network provides for organisms in terms of their evolvability. Additionally, the informative signals in NNs, far from representing a weak linkage, connect to hundreds or thousands of other entities in the network through fine-tuned weights, exemplifying what can be called a \textit{strong linkage}. This means that even if, hypothetically, one could decouple multiple learned substructures from different neural networks, they can’t be easily recombined into a new process that, in theory, \textit{should} be possible—because the connecting weights of the original processes have been fine-tuned specifically for their respective networks.\footnote{To exemplify: Suppose two sequential processes, each corresponding to a portion of a neural network, where the output of one becomes the input of the other: $A(.)$ (e.g. detection of some object) and $B(.)$ (e.g. a policy to move to where the object is) form our goal process $C=B(A)$. As the simplest case, suppose that we learn $C$ with two neural networks, in form of $C_1=B_1(A_1)$ and $C_2=B_2(A_2)$. In such a case, recombining them to form $C_3 = B_2(A_1)$ would not be functional. This is because the output of $A_1$ would have been fine-tuned to the input weights of $B_1$, not $B_2$. For example, if the corresponding input weights of $B_1$ are twice those of $B_2$, the output of $A_1$ will be half the magnitude required to activate $B_2$ at the same level as when using $A_2$.}

\subsubsection{Drivers of recursively-accelerating evolution}

A large number of fundamental biological processes—including both the basic processes responsible for generating fundamental structures (\citep{holstein2012evolution}) or higher-order traits (\citep{lemons2006genomic}), as well as processes that allow organisms to generate adaptive responses (\citep{holstein2012evolution})—evolved early in evolutionary history and have been mostly preserved since then, being shared across distantly related organisms (\citep{marc2005plausibility, gerhart2007theory}). Major evolutionary changes since these early developments have primarily occurred at the regulation of these \textit{core processes}. In this respect, an evolutionary theory that incorporates developmental processes offers a better explanation for the observed accelerating rate of change in phenotypic complexity than Modern Synthesis alone: Once core processes evolve, they can be reused with minimal genetic changes that focus on regulating these processes (\citep{marc2005plausibility, gerhart2007theory}), and thus be recombined into new core processes that serve as an even more competent basis for further evolution, and so on.

This qualitative pattern of evolutionary progression, as explained by developmental and regulatory principles, is entirely different from the progression of learning seen in contemporary machine learning. At the same time, it reflects the characteristics of an ideal learning progression that artificial learning systems strive to achieve: Core functionalities that help an agent throughout its lifetime and across tasks are learned early on, possibly through a long training procedure, and then can be reused, recombined, and form the basis for further learning, all without destroying existing structures, thereby enabling open-ended and unbounded learning. As such, it allows for a recursively accelerating pace of acquiring new knowledge, where existing knowledge facilitates future learning both by serving as more competent building blocks and by aiding in the information acquisition process.

\subsubsection{\textbf{Adaptation via local variation and selection}}
\label{sec:edb_process_varsel}

\textit{Exploratory processes} in organisms (\citep{gerhart2007theory, marc2005plausibility, west2003developmental}) leverage the \textit{variation and selection} principles underlying biological evolution somatically and locally, where variants of a substructure are generated and then refined through a selective signal. This allows organisms to respond to the needs and circumstances of either the environment or the rest of their bodies on-demand. The operation of the immune system, which relies on the selective proliferation of lymphocytes recognizing antigens with introduced mutations to enhance specificity, provides a prominent example of adaptive somatic response by both generating an immediate reaction and offering long-term immunity (\citep{burnet1957modification, burnet1957modification, rajewsky1996clonal}). Similar principles govern neural development (\citep{hiesinger2021self, marc2005plausibility}), where synapses \& axons are initially overproduced and later refined through pruning of weaker ones (\citep{bourgeois1989synaptogenesis, rakic1986concurrent, chechik1998synaptic, sakai2020synaptic, petanjek2023dendritic, lamantia1990axon, rakic1983overproduction, provis1985human, kaiser2009simple, innocenti1997exuberant}). As such, local somatic variation and selection serve as the primary mechanisms for generating adaptive responses and structures in situations where no evolutionary preprogramming regarding their ultimate form is possible. Additionally (and crucially), these processes enable learning without strongly informative signals—like the gradient signals in neural networks—thereby allowing the acquisition of structures whose components are connected by weak linkage (see above).

The presence of variation and selection as a means for in-system adaptation (as opposed to an external evolutionary process operating on the variants of the complete system, like population-level evolution) is significant due to some important properties of this process: Variation and selection enable adaptation to new circumstances while preserving consistency with past circumstances, a capability closely related to the goal of designing AI agents that can learn continually without destroying past knowledge. In particular, the challenge of learning or adapting without erasing past knowledge/adaptations, even when the pressure toward their selection is alleviated, pertains to a fundamental uncertainty that any agent (biological or artificial) faces regarding the future: For any given observation or circumstance encountered at a specific moment in time, the agent can generate multiple plausible solutions or internally consistent models/explanations. Among these alternative solutions, many may be \textit{neutral} with respect to each other (\citep{wilke2001adaptive, tenaillon2020impact}), meaning they perform similarly well in the present context but differ in their suitability for future scenarios. Since an agent cannot \textit{a priori} determine which of these neutral alternatives will prove most beneficial in the long run, the natural resolution to this problem is a mechanism of variation and selection—multiple solutions of comparable immediate utility are generated and explored in parallel (\citep{wagner2011origins}), and over time, those that prove most effective in future contexts are retained, while the rest are discarded. This process, in the absence of other pressures (such as natural decay via random changes or pressure towards simplification due to a fitness cost in complexity), allows for adaptation to new circumstances without compromising the agent's adaptation to past environments/observations. This principle is well-understood in population-level evolution (\citep{wagner2011origins}); however, we are not aware of a somatic exploratory processes that exemplifies this specific property, or has been analyzed from this perspective. Nevertheless, we demonstrate an example of this principle in computational design in Section \ref{sec:Modelleyen}, showing that a particular form of local variation and selection can indeed be the key to realizing continual learning in AI systems.

This brings us back to our discussion of contemporary machine learning. The reader may think that local variation-selection is also a mechanism that has not found its way into the AI of today, but in fact, it has: Even though it was not intentionally designed as such, overparameterized neural networks trained through gradient descent can be seen as undergoing a variation and selection process. Initially, there is an abundant \textit{pool of variation} (effective functionality encoded by the combination of multiple connected parameters in the network, i.e., "weight patterns," which are abundant at the start due to random initialization of the network). This variation is then crushed by the selective force of gradient descent, amplifying "beneficial" weight patterns that contribute to reducing error while diminishing others. This process converges to a minimum where the remaining weight patterns are non-random and functional. The problem here is not the absence of the mechanism, but rather its limited nature: This variation-selection process is \textit{not iterative}; there is no mechanism for regenerating variation atop existing structures. Instead, it is a singular selection process that transforms the large, initial variation pool into a well-adapted network. New variation cannot be introduced over this adapted structure without disrupting existing knowledge, as any new selective pressure (e.g., a gradient in a new direction) affects the entire network. This perspective not only offers a new interpretation of the expressive power of neural networks—an immense initial reservoir of variation refined by a strong selective signal—but also provides a new insight into the destructive adaptation problem in continual learning: Fundamentally, the issue arises from the network’s inability to generate new variation locally when and where needed, as is indispensably required for further learning, without annihilating prior weight patterns.. This contrasts sharply with exploratory processes in biological systems that can locally generate variation on demand.

\subsection{Outlining a new design philosophy for AI}
\label{sec:ai_future}

The core proximal principle that explains the explanatory gaps in the Modern Synthesis regarding the observed progression pattern of evolution, which also finds its correspondence in contemporary machine learning (namely, rapid evolvability, recursively accelerating pace of adaptation, and an open-ended, unbounded potential for further evolution — as shown in the second column of Table \ref{tab:ms_ml_analogy}), seems to be summarized as the structural organization of the organism at the phenotype level (first row of Table \ref{tab:ms_ml_analogy}). This structural organization provides encapsulation and facilitates adaptive reuse. As discussed in Section \ref{sec:contemporary_ai}, these potential benefits of structural organization for AI systems are already widely recognized as desirable traits. Yet, despite their intuitive appeal, surprisingly little practical progress has been made, largely due to the rigidity and predefined nature of such structures (\citep{su2024focuslearn, goyal2020object, pateria2021hierarchical}), or their inherent connection to other limitations of neural networks (\citep{andreas2017modular, devin2017learning, sahni2017learning, goyal2019reinforcement, yang2020multi, pateria2021hierarchical, iyer2022avoiding}). Achieving the same level of structural organization across all levels of system organization seems to be a prospect that does not even seem relevant to current AI systems.

The critical insight here from EDB lies in \textit{developmental principles} which, as outlined above, govern the structure of biological organisms and underlie the recursively-increasing pace of evolution. While structural organization and adaptive reuse are desirable as \textit{sub-means} toward properties like system-level continual learning and comprehensibility, they are not the \textit{fundamental means} themselves. The fundamental means of adaptability, on the other hand, are rooted in the principles of low-level developmental mechanisms. AI research must recognize this and shift its focus accordingly, embracing a new design paradigm grounded in these principles. To summarize our discussion above, we highlight the following key points that are directly relevant to the current state of AI (each of which is concretely exemplified by the designs described in Section \ref{sec:demos}):

\begin{itemize}
    \item \textit{Regulability:} (Section \ref{sec:dirad}) The capability for genetic regulation forms the foundation of all structural properties in complex organisms, including the reuse of processes, modular and hierarchical organization. It furthermore enables innovation, complexification, and differentiation through changes that are initially neutral and non-detrimental, allowing organisms to surpass previous local adaptive peaks, which are characterized by regions corresponding to learned parameters representing "intermediately-optimal" values, as defined by statistical trade-offs across samples (see Section \ref{sec:contemporary_ai}). In machine learning systems, the "learned structures" typically refer to the connections between computational units (neurons). To replicate these properties in AI, a redesign of the approach must include the possibility of regulating these connections themselves.

    \item \textit{Weak linkage:} (Section \ref{sec:Modelleyen}) Weak linkage allows for the convenient reuse and recombination of core processes learned previously, while also preserving the original form of the process by abstracting its internal details into a simple input-response relationship. In contrast, contemporary machine learning exemplifies an extreme case of what can be called strong linkage, with densely interconnected and fine-tuned units — an approach that must be revised if the versatility of recombination found in biological systems is to be achieved. It is worth noting that weak linkage is not a principle exclusive to biological systems, but is also foundational in many engineering disciplines (crucially, outside today’s AI), where similar properties—such as ease of reuse, recombination, and decoupling—are prioritized.

    \item \textit{Learning via component-level variation and selection:} (Section \ref{sec:Modelleyen}) Variation and selection, applied somatically at the component level, serve as the primary means of adaptation in organisms when no evolutionary preprogramming is possible, preserving previous adaptations while allowing for adaptation to new circumstances—uniquely across all the alternative principles of adaptation or learning that we are aware of. Large neural networks, initialized randomly and learned through a gradient signal, correspond to a single iteration of a variation-selection cycle, where an initial pool of variation is refined into a well-adapted response through a strong selective signal, but without any means of regenerating variation when needed, providing one of the reasons behind the destructive adaptation problem. Therefore, the revised machine learning design paradigm must incorporate mechanisms for generating structural variation locally, where and when needed, to form the foundation of future adaptability, and leverage the iterative process a way that the properties of this cycle can be guaranteed to preserve previously learned knowledge. Importantly, as will also be evident in Section \ref{sec:Modelleyen} (and as seen in biological systems, such as synaptic variation/selection or the operation of the immune system mentioned above), local variation–selection serves as a mechanism for learning adaptive relationships between components connected via weak linkage—precisely in contexts where no strong, fine-tuned, informative signals exist, such as the gradient signals used in neural networks.
 
\end{itemize}

Without developmental principles operating at the foundational level to generate structure rather than imposing it from above, with processes that realize encapsulation or reuse coming only on top of a system that already learns and organizes itself by these principles, it becomes impossible to achieve the desired structural properties in learning systems \textit{adaptively} and \textit{across all levels of organization}, the latter being particularly necessary for unbounded recursive improvability. A new design paradigm that views these properties — a desired pattern of learning progression, a well-structured organization, and developmental principles — as an integrated whole—rather than as properties that can be realized in isolation—is therefore essential.

Complementarily, the complexity of core control processes in biological systems indicates that the appropriate design level for adaptive systems is \textit{not} high-level integration, but rather \textit{more proficient low-level organizational units}. This stands in contrast to the majority of current ML research, which focuses primarily on high-level mechanisms operating on networks of perceptrons (\citep{wan2024towards, colelough2025neuro}) or building block alternatives with qualitatively similar capabilities (\citep{bal2024rethinking}), with limited interest in fundamentally redesigning the capabilities of these core building blocks. It is unlikely that the design principles outlined above can be applied across all levels of organization using fundamental entities as simple as perceptrons, underscoring the need for research in redesigning our core computational units to align with the developmental capabilities as discussed previously. Note that more capable computational building blocks will likely eliminate the necessity for complex integration mechanisms layered atop the fundamental learning algorithm, as their objectives will now be achieved by the low-level units (see the handling of preservation of past knowledge in Section \ref{sec:Modelleyen})—this should hence be viewed as a shift in the relevant design level (toward lower levels) rather than the introduction of new design goals.

Finally, we must discuss a perceptual capability historically regarded as a backbone of human intellect, yet conspicuously absent in contemporary AI: the capability for \textit{abstraction}. One way to interpret the insights from EDB is as an explanation for the generation of the organism’s internal multi-level \textit{organizational structure}\footnote{Of course, this is also intricately related to multi-level selection theories (\citep{damuth1988alternative, Okasha2005multilevel, gardner2015genetical, czegel2019multilevel}) and evolutionary transitions (\citep{smith1997major}), yet we omit them from our discussion as they do not directly pertain to the focus of this paper.}. When applied to AI, this concept of a multi-level structure—characterized by \textit{hierarchy, modularity, and reuse of common substructures} as discussed above—can be seen as a method for generating \textit{abstract} representations at multiple scales. These representations might include high-level percepts (e.g. features and objects), equivalence sets (e.g. percepts with shared outcome or alternative outcomes of the same percept), or behavioral patterns (e.g. subpolicies (\citep{bakker2004hierarchical, li2019sub})). In addition to their potential for introducing comprehensibility to learned models (conditioned on appropriate analysis and interpretation of their internal composition, just like any hierarchically-organized software), such abstract representations can enable a more separable and precise depiction of the knowledge needed for high-level cognitive processes (i.e. non-learning processes often categorized as "classic" or "symbolic AI" today, such as goal-oriented behavior) that are notoriously difficult to integrate into modern learning systems that function as black boxes between raw observations and low-level actions/outcomes.\footnote{This integration is sought in current design paradigm by fields like neurosymbolic AI (\citep{wan2024towards,colelough2025neuro}) or forward-sampling based methods to perform goal-oriented behavior (\citep{mcmahon2022survey, otte2015survey}). Such attempts, however, are limited by the analogues of the same general constraints of contemporary ML systems as outlined in Section \ref{sec:contemporary_ai}, which we do not detail again to avoid repetition.} As such, a new design philosophy—grounded in the conceptual analysis laid out in this section—can serve as a critical step toward equipping intelligent agents with the essential capability to generate abstract representations expected to have many important downstream ramifications.

A design paradigm that incorporates insights from evolutionary developmental biology, hence, holds immense potential in resolving multiple major inherent limitations of contemporary machine learning, \textit{simultaneously} and in an \textit{organic} manner. Research in artificial learning systems today has neither a widespread, generally applicable solution to any of these problems, nor a recognized research agenda grounded in solid theoretical foundations to address them. Evolutionary theory, as augmented by developmental principles, provides a unified conceptual foundation that can plausibly usher in a new design philosophy and paradigm for both learning systems and artificial intelligence in general.human

\section{Design principles from EDB in action: Two demonstrations for learning systems}
\label{sec:demos}

In the previous section, we outlined the conceptual basis that EDB can provide for the future of AI and how it reflects in-principle to design principles for AI. In the following two sections, we briefly discuss two example designs for learning systems. These demonstrations exemplify the incorporation of the aforementioned principles into learning systems and address some major limitations of contemporary ML with theoretical guarantees. Specifically, the first design (Section \ref{sec:dirad}) demonstrates how the need for overparameterization to find a good solution is overcome through regulatory mechanisms, while the second design (Section \ref{sec:Modelleyen}) shows how the loss of past knowledge can be prevented by combining a learning mechanism based on a local variation-selection process with weakly linked input and target relations. Furthermore, both designs provide new insights into how these corresponding processes can facilitate evolution (with respect to their relative benefits) and somatic adaptation in biological systems as well. 

We note that these demonstrations follow from our main argument above and aim primarily to illustrate and exemplify the points we’ve outlined. They are \textit{not} required for fully understanding our conceptual argument and can be skipped by readers without a background in contemporary AI approaches. However, they are useful for seeing these principles in action. To maintain focus on our main arguments, we do not delve into extensive technical discussions or experimental validations of the described designs but instead outline the key components of the designs related to evolutionary principles and their relevant properties. Interested readers can find detailed information and experimental validations in \citep{erden2024directed} for Section \ref{sec:dirad} and in \citep{Erden2024Modelleyen, erden2025agential_extendedabs} for Section \ref{sec:Modelleyen}.

\subsection{Demonstration 1: Complexification via regulation resolves conflicting pressures and guarantees a good solution}
\label{sec:dirad}

The first demonstrative design (coded "D1") is a growth method for continuously parameterized networks learning by gradient descent (\citep{dai2020incremental, dai2019nest, maile2022and, evci2022gradmax, maile2023neural, maile2022structural}), which achieves minimally extensive complexification by forming points of regulation as new nodes in place of previously existing edges where statistical trade-offs (gradients in opposite directions) are present. This method of complexification, as shown below, guarantees that a network can achieve a very good solution without the need for large, overparameterized networks. While the reader, if they want, can interpret this design as a growing "neural network" for simplicity, we refrain from using this term in this section, as the mechanisms described do not appear to have a plausible neurobiological correspondence to our knowledge.

Consider a network whose learnable parameters (weights of edges and biases of nodes) are updated with a gradient that represents the average influence of multiple examples, as is the case in neural networks (Eq. \ref{eq:nn_grad_update}). In Section \ref{sec:contemporary_ai}, we had discussed how such an approach is prone to getting stuck in a local minimum of the cost function (analogous to a local adaptive peak in a fitness landscape) due to opposing pressures from different examples on a given parameter. Let’s formalize this concept with the following definition.

\begin{definition}
   (Adaptive potentials (APs)) The \textbf{immediate adaptive potential (AP)} of a
   parameter $\alpha$ is defined as the net gradient that this parameter gets over a given batch, $\partial C/\partial \alpha$ where $C$ is the total cost (error). The \textbf{total AP} of $\alpha$ is defined as $\sum_{m \in D}|\partial C^m/\partial \alpha|$ (where $C^m$ is the cost associated with an individual sample $m$ and $D$ is the whole training data), as a measure of the total adaptive gain that can be obtained by a change in $\alpha$, if it could be exploited. We define that the immediate (total) adaptive potential of $\alpha$ is \textbf{exhausted} if immediate (total) AP is $\approx 0$.
\end{definition}

\begin{figure}
     \centering
     \begin{subfigure}[t]{0.32\textwidth}
         \centering
         \includegraphics[width=\textwidth]{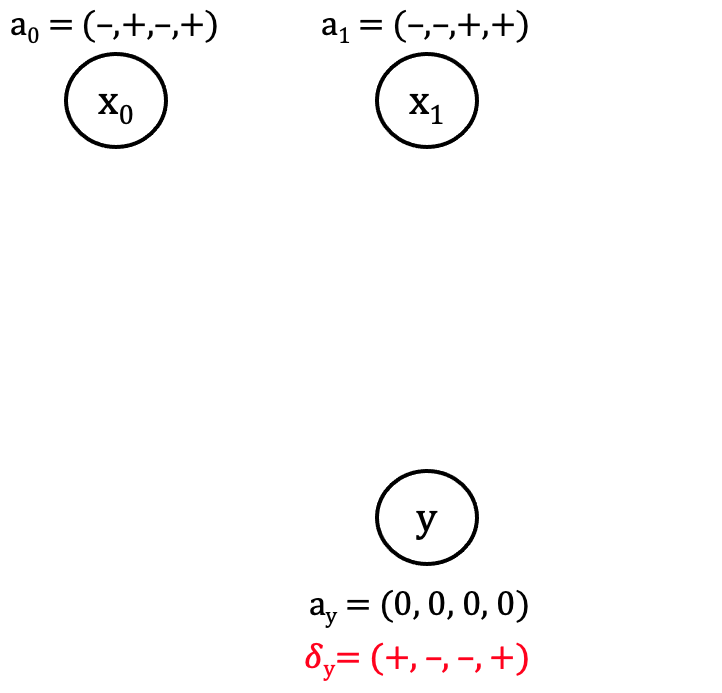}
         \caption{}
         \label{fig:dirad_example_0}
     \end{subfigure}
     \hfill
     \begin{subfigure}[t]{0.32\textwidth}
         \centering
         \includegraphics[width=\textwidth]{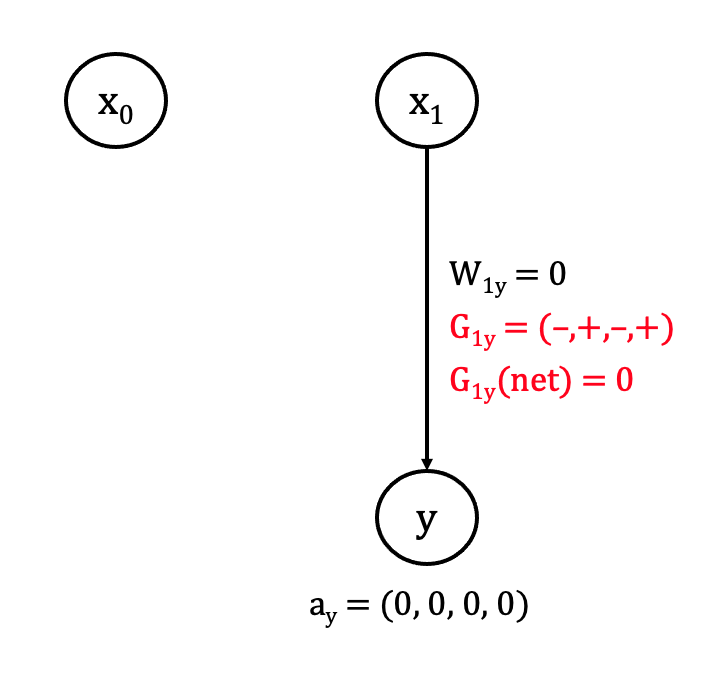}
         \caption{}
         \label{fig:dirad_example_1}
     \end{subfigure}
     \hfill
     \begin{subfigure}[t]{0.32\textwidth}
         \centering
         \includegraphics[width=\textwidth]{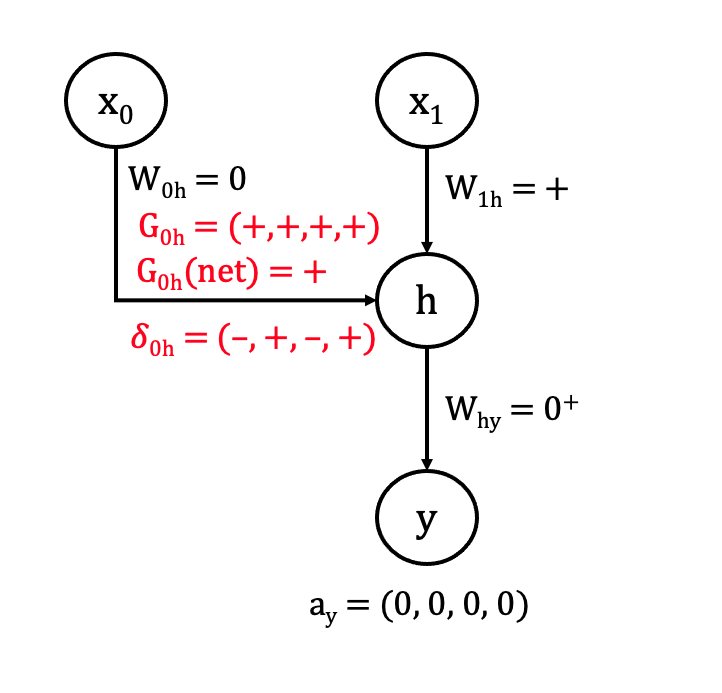}
         \caption{}
         \label{fig:dirad_example_2_3}
     \end{subfigure}
     \hfill
     \begin{subfigure}[t]{0.32\textwidth}
         \centering
         \includegraphics[width=\textwidth]{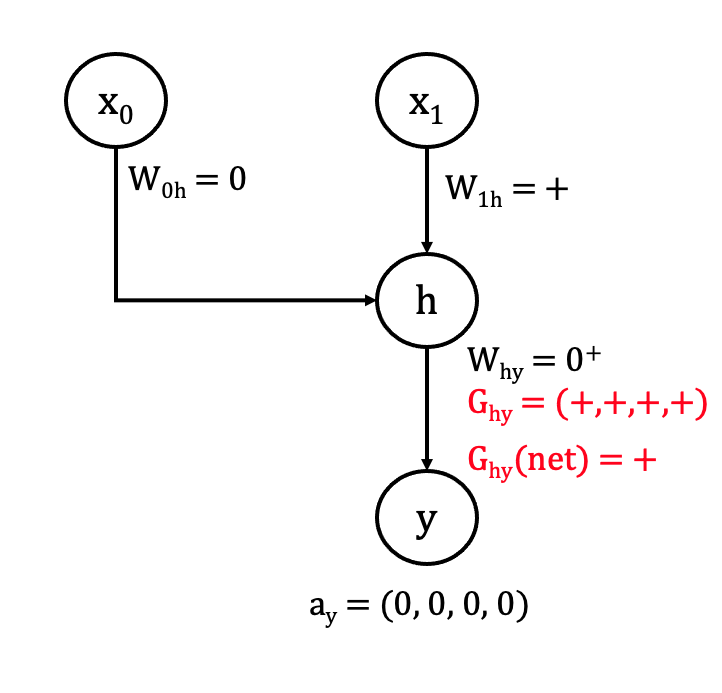}
         \caption{}
         \label{fig:dirad_example_4}
     \end{subfigure}
     \hspace{1cm}
     \begin{subfigure}[t]{0.32\textwidth}
         \centering
         \includegraphics[width=\textwidth]{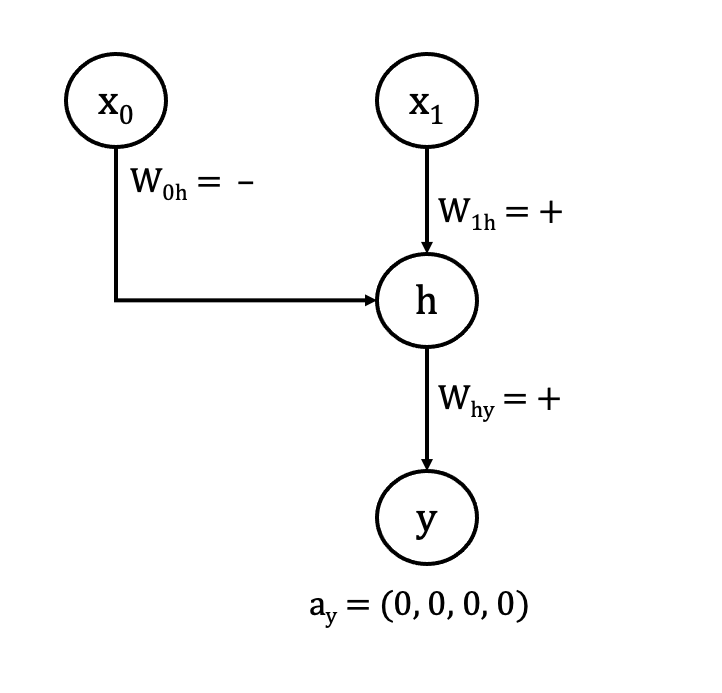}
         \caption{}
         \label{fig:dirad_example_5}
     \end{subfigure}
    \caption{A simplified illustrative case of the path of adaptation for signed XOR for D1 ("False" represented by $-1$ instead of $0$). Inputs: $x_0$, $x_1$. Output: $y$. In the figures, $G_e$ represent $dC/dw_e$, $\delta_n$ represent $dC/dz_n$, $a_i$ state of node $i$, and the four values in parentheses represent the signs that a variable takes for the four samples, respectively. We simplify the illustration by assuming no bias and that the adaptation process of different components happen in sequence instead of simultaneously. \textit{(a)} Initial state. $y$ has immediate AP exhausted but total AP nonzero, hence will form an edge. Neither input is an immediately-useful source, since neither matches with gradients of $y$. \textit{(b)} $y$ forms an in-edge, source chosen randomly. The generated edge has a net gradient of 0 and hence cannot proceed with adaptation. This is a "local minimum" in a static network. \textit{(c)} The new edge undergoes ENC from exhaustion, and its gradient is transferred to the Term 1 gradients of the new node $h$. $h$, under immediate AP exhaustion, gets an edge from $x_0$ that can provide perfect match with the sign of its gradients and creates a large positive net gradient for the new edge. \textit{(d)} Modulatory edge to term 1 of the converted node $h$ stabilized in a negative value following adaptation. The net modulatory effect of $x_0$ on the $x_1 \rightarrow y$ pathway inverts the sign of gradients when $x_0$ is positive. The net gradient of $w_{hy}$, previously 0, goes net positive as a result of the modulation. \textit{(e)} Final stable state with correct response in $y$.}
    \label{fig:dirad_example}
\end{figure}

The phenomenon of being stuck in a local minimum due to a \textit{statistical trade-off} in a parameter arises when the immediate AP on a parameter is exhausted (no net gradient), while its total AP is not (still large pressure exerted by at least some examples) — representing a potential that cannot be utilized, regardless of no matter how large, since the net gradient is zero. This issue can be overcome by the possibility of regulation at the edges where this phenomenon occurs, through a network that starts with only the input and output nodes (Figure \ref{fig:dirad_example_0}) and complexifies via two \textit{neutral} generative processes (neutral in the sense that they do not change the immediate network responses \citep{wagner2011origins} — see \citep{erden2024directed} for the demonstrations of their neutrality):

\paragraph{1. Edge generation} This process (Figure~\ref{fig:dirad_example_1}) generates an in-edge with an initial weight of $0$ to a node $j$ if the immediate AP of the node's bias is exhausted \textit{and} the immediate AP of all its in-edge weights are exhausted as well, while the total AP of the activation input to the node ($z$ in Eq.~\ref{eq:nn_y}) remains non-exhausted (Figure~\ref{fig:dirad_example_1}). The source of the edge is chosen from the candidates to be the one that maximizes the magnitude of the expected immediate gradient update on the edge (i.e., the value $\left| \partial C/\partial w_{ij} \right| \propto \left| \sum_m a^m_i (\partial C/\partial z^m_j) \right|$, where $a^m_i$ is the state/response of node $i$, and the right-hand side is derived via the chain rule applied to Eq. \ref{eq:nn_y}). Intuitively, this operation allows us to relieve a node with nonzero total adaptive potential from the exhaustion of its immediate AP, provided there are sources that align well with the change directions indicated by the gradients.

\paragraph{2. Edge-node conversion (ENC)}

The ENC mechanism corresponds to the formation of a \textit{point of regulation} at an edge where the immediate AP is exhausted, but its total AP is not. In such a situation, ENC resolves the exhaustion of immediate AP by modulating the gradients of the original edge (upon the progression of adaptation) so that they become aligned instead of opposing to one another. Formally, when an edge $(i,j)$ undergoes ENC, it is replaced with a new node $k$ and two edges $(i,k)$ and $(k,j)$ that become the new path connecting $i$ to $j$  (Figure \ref{fig:dirad_example_2_3}). The new node is \textit{modulatory}, whose state is computed by the multiplication of two terms:

\begin{equation}
\label{eq:sec2modnode}
    \begin{split}
        a^m_x = \prod_{i \in \{0,1\}}\sigma_i(\sum_{y \in in_i(x)} w_{yx}a^m_y + b_{x,i} )
    \end{split}
\end{equation}

where subscripts ${x,i}$ refer to \textit{node x, term $i$}. Two distinct transfer functions are used for the two terms, where $\sigma_0(z)=z$ and $\sigma_1(x) = 4/(1+e^{-Kx}) - 1$, $K$ being a constant. Bias of the first term is set to a fixed $0$, and the weights of new edges are set as $w_{ik}=1$ and $w_{kj}=w_{ij}$. These design decisions serve to ensure the neutrality of the operation (see \citep{erden2024directed}) and are not necessarily unique choices in this regard. Furthermore, the chosen function for $\sigma_1(\cdot)$ can take values in the range $(-1,3)$, and hence is able to invert the sign of the previously opposing gradients. The two terms in the modulatory nodes will have distinct gradients, $\partial C/ \partial z_{x,i}$, and will form in-edges separately for these two terms, as per the edge generation process described above. At the time of ENC, the original source $i$ is connected to term 0 of the new node $k$, and no node is connected to term 1.

It can be shown that immediately after this conversion, the gradients for the activation input of term 1 of the new node are equal to the gradients of the original edge, $\partial C/\partial z^m_{k,1} = \partial C/\partial w_{ij}$. Through ENC, one effectively transfers the opposing per-sample weight gradients of the original edge (which cannot be treated as a vector for adaptive purposes, but only in terms of their average effect, which cancel each other) to the deltas of a node (which can be treated as a vector that can create an adaptive change even if their average is 0). If a proper source $l$ can be found for term 1 of node $k$ that yields a nonzero $\partial C/\partial w_{lk}$, then the net gradient of the new edge $(k,j)$ will become nonzero as the state of term 1 of node $k$ is adapted under the influence of gradients proportional to those of the original edge, thus escaping what would have been a local optimum in a static network (Figures \ref{fig:dirad_example_2_3} and \ref{fig:dirad_example_4}).

\begin{figure}
     \centering
     \begin{subfigure}[t]{0.47\textwidth}
         \centering
         \includegraphics[width=\textwidth]{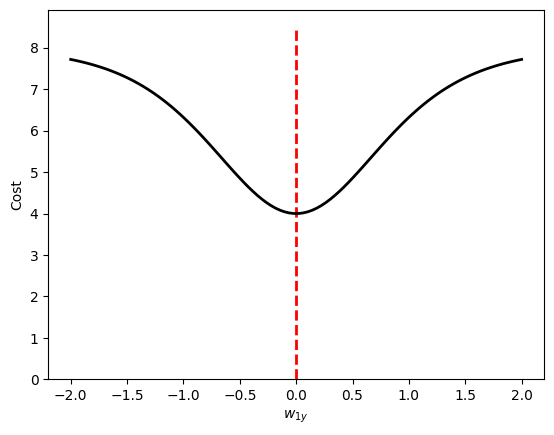}
         \caption{Before ENC (Fig. \ref{fig:dirad_example_1}).}
         \label{fig:dirad_landscape_pre}
     \end{subfigure}
     \vline
     \begin{subfigure}[t]{0.50\textwidth}
         \centering
         \includegraphics[width=\textwidth]{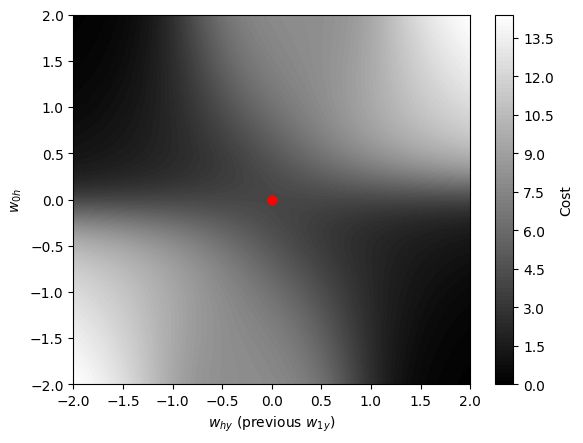}
         \caption{After ENC (Fig. \ref{fig:dirad_example_2_3}).}
         \label{fig:dirad_landscape_post}
     \end{subfigure}

    \caption{The cost function landscape before and after ENC during XOR learning (Fig. \ref{fig:dirad_example}), with $K = 1$, and high and low targets encoded as $-1$ and $+1$ respectively, and outputs transformed by the  $\tanh(\cdot)$ activation function. It is clearly observed that when $w_{1y} = 0$, the point of immediate AP exhaustion that triggers ENC is a local minimum (a). However, ENC and the subsequent formation of a "regulatory" connection reconstruct the cost function landscape with an extra dimension (b), transforming the point corresponding to $w_{1y} = 0$ (shown with a red dot) from a local minimum into a saddle point, from which a smooth high-gradient path toward either of the two perfect zero-cost solutions (top-left or bottom-right) can be initiated with a small deviation. Note that for more complex functions and networks, this process continues across higher-dimensional spaces as long as there is still latent adaptive potential in the network.}
    \label{fig:dirad_landscape}
\end{figure}

To rephrase this verbally, the regulatory connection formed by the ENC operation effectively redefines the cost function in a new parameter space with an additional dimension, such that the point in the original parameter space corresponding to the current system configuration is transformed from a local minimum into a saddle point. This transformation introduces a nonzero gradient in the vicinity of the point, as shown in Fig. \ref{fig:dirad_landscape}. This parallels, and provides a formal mathematical perspective on, the role of regulatory mechanisms in biology that enable organisms to go beyond points that would otherwise be local adaptive peaks (\citep{west2003developmental, wagner2011origins, moczek2011role, poelwijk2007empirical}), particularly those that can be interpreted as the selective expression of conflicting, alternative responses (as discussed in Section \ref{sec:edb_structure_genereg}). Notice also that the newly generated edge is no different from any other edge in the network and can undergo further ENC if and when necessary.

These regulatory connections not only resolve statistical trade-offs locally but also guarantee that the adaptation process will continue (theoretically, conditioned on resource availability and numeric sensitivity limitations in computational experiments) until a very strong condition is met:

\begin{theorem}
    A chain of ENC operations will propagate, ultimately yielding a nonzero net gradient in at least one of the formed edges, thereby enabling adaptation to proceed throughout the network, until the following condition is satisfied:

        \begin{equation}
        Cov(\prod_{x\in A}a^m_x, \frac{\partial C^m}{\partial w_{ij}}) = 0, \ \forall A \in P(N)
        \end{equation}

    where $N$ is the set of all available candidate sources for edge formation (always covering all input nodes) and $P(N)$ its power set.
\end{theorem}

See \citep{erden2024directed} for the proof, which follows the propagation of the condition for a net gradient to be zero across all possible connections that can be formed in the network. In the most relaxed case, this condition states that adaptation will proceed as long as there is any nonzero correlation between the gradient vector of the edge we aim to take out of AP-exhaustion and any of the potential multiplicative combinations of the input nodes in the network. This is a very strict condition, far beyond simply having a mean $\frac{\partial C^m}{\partial w_{ij}} = 0$, as would be the case in static networks. We intuitively suspect, though did not verify mathematically, that this corresponds to a global optimum guarantee.

\paragraph{Significance}

Design D1 leverages the principle of regulatory connections (specifically to the subjects of evolution/learning, i.e., the edge weights in continuous networks) to guarantee convergence (via continuation of adaptive operations in direction of gradient) to a good solution, characterized by a very strict condition that we suspect to be a global optimum. This is achieved by the added regulatory connection altering the fitness landscape itself, in a way that creates a high immediate gradient where there once was a local minimum, while also enabling further upstream regulability of connections that were themselves initially regulatory. Additionally, this approach provides valuable insight into the role of regulatory mechanisms in the evolutionary process, particularly in how they influence the fitness landscapes around regions otherwise marked by local adaptive peaks; and can be viewed as a model of differentiation into alternative phenotypes, which can then follow their own adaptive pathways, advancing along the fitness gradient in a manner impossible without the isolation and differentiation of these two contexts (due to opposing effects making the system be stuck in a local minimum) \citep{vaishnav2022evolution, west2003developmental}.

\subsection{Demonstration 2: Local variation-selection and weak linkage enables continual learning without destructive adaptation}
\label{sec:Modelleyen}

In this second demonstration ("D2"), we illustrate how the principle of learning through local, component-level variation and selection, using weakly-linked information sources, leads to an approach inherently capable of continually learning new information without erasing past knowledge. Additionally, this process occurs rapidly, avoiding the lengthy, repetitive passes through training data typically seen in traditional statistical learning methods. We’ve already discussed the unique potential of a variation-selection process to enable adaptation to new scenarios while retaining adaptability to previous environments (as exemplified by its unfolding at population-scale evolution) in Section \ref{sec:edb_process_varsel}. It turns out that a simple implementation of this principle, combined with the insights from regulatory connections discussed in D1, can realize such a learning progression in a theoretically unbounded manner. As before, we only outline the core concepts here in relation to our discussion, and for full details, we direct the reader to \citep{Erden2024Modelleyen, erden2025agential_extendedabs}.

This design, unlike neural networks or the approach in D1, does not operate with continuously parameterized values but rather with discrete state variables. This constraint is not overly limiting, as many AI tasks can be described using, or translated into, non-continuous variables or structures (for example, network representations of complex visual spaces (\citep{erden2025mnr})). The only potential exception we can think of is low-level, fine-tuned control tasks, which generally fall more within the domain of control theory than artificial intelligence. A \textit{state variable (SV}) in this system is this defined as a variable that can take an \textit{active}, \textit{inactive} or \textit{unobserved} value, represented as $1$, $-1$, and $0$ respectively. These SVs may either be internal components of the model learned by the system or externally defined, either as information sources (inputs) or as prediction targets. A specific type of internal state variable in this system is the \textit{conditioning state variable (CSV)}, which represents learned relationships across other types of SVs in the system (e.g., inputs and outputs to be predicted) and as such, also serve as the locus of learning processes.

\begin{definition}
    (Conditioning SV - CSV) A CSV $C$ is a type of SV with mutable sets of positive sources $X_P$, negative sources $X_N$, and conditioning targets $Y$, all of whom are other SVs in the model. The sources of $C$ are considered \textit{satisfied} if all positive sources are active and all negative sources are not active. If sources are satisfied, state of $C$ is \textit{active} if sources are satisfied and the targets are active, inactive if sources are satisfied and targets are \textit{inactive}, and \textit{unobserved} otherwise.\footnote{This is a simplified version of the full definition of CSVs as presented in \citep{Erden2024Modelleyen, erden2025agential_extendedabs}, where certain properties irrelevant to our discussion have been omitted to avoid unnecessary complication. For complete details, please refer to the cited papers.}
\end{definition}

CSVs model the presence or absence of a relationship between a learned condition (sources) and its effect (active target states), with the CSV being active (present) or inactive (absent) to indicate this relationship. Note that CSVs are \textit{not} feedforward computational units; they represent the relationship between sources and targets - states of their target states are set independently of the CSV (e.g. externally defined output states), unlike feedforward units that determine target states based on source states. CSVs partially function as feedforward units only when used for prediction of alternative outcomes.

\begin{figure}
     \centering
     \begin{subfigure}[t]{0.40\textwidth}
         \centering
         \includegraphics[width=\textwidth]{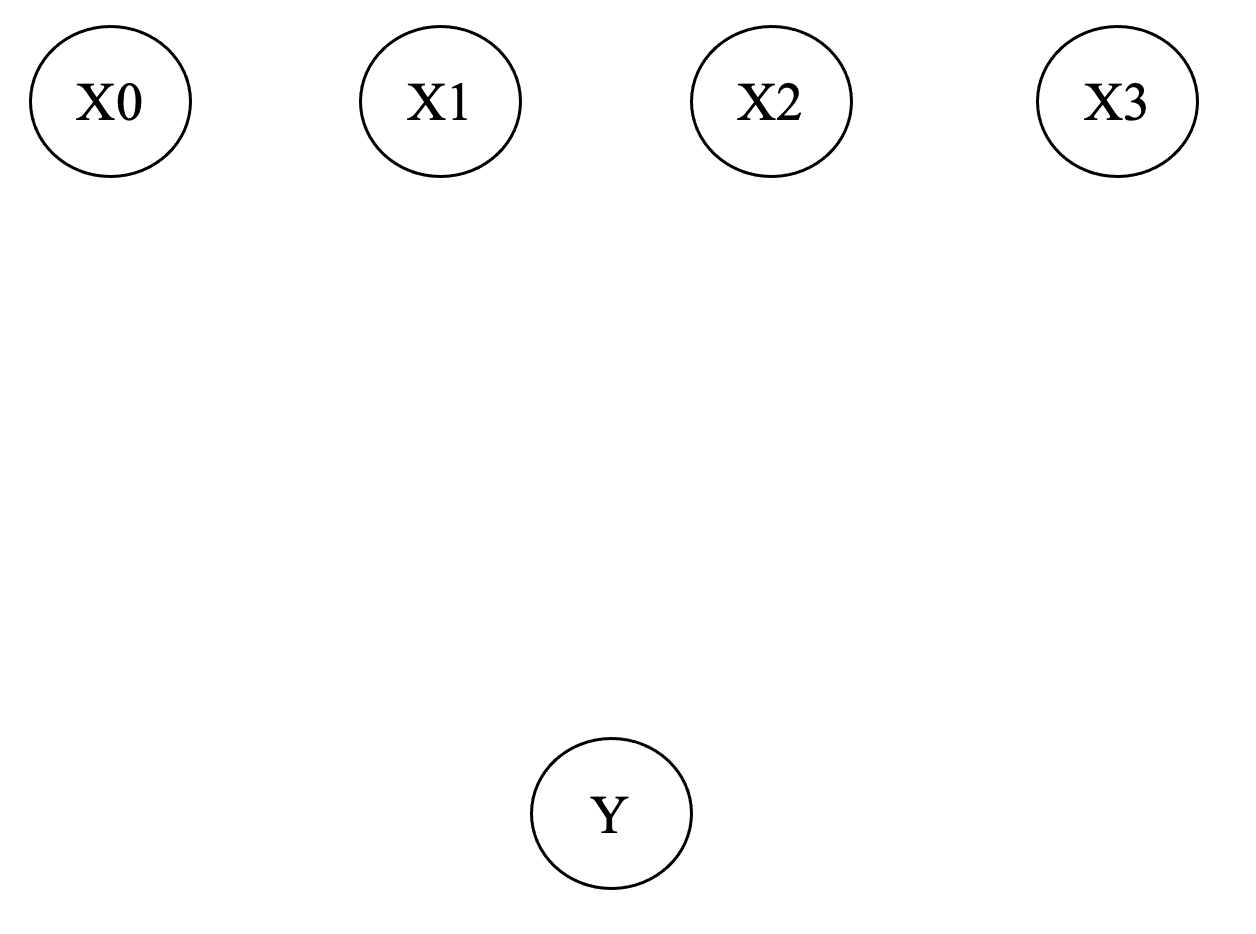}
         \caption{}
         \label{fig:csvform_1}
     \end{subfigure}
     \hfill
     \begin{subfigure}[t]{0.40\textwidth}
         \centering
         \includegraphics[width=\textwidth]{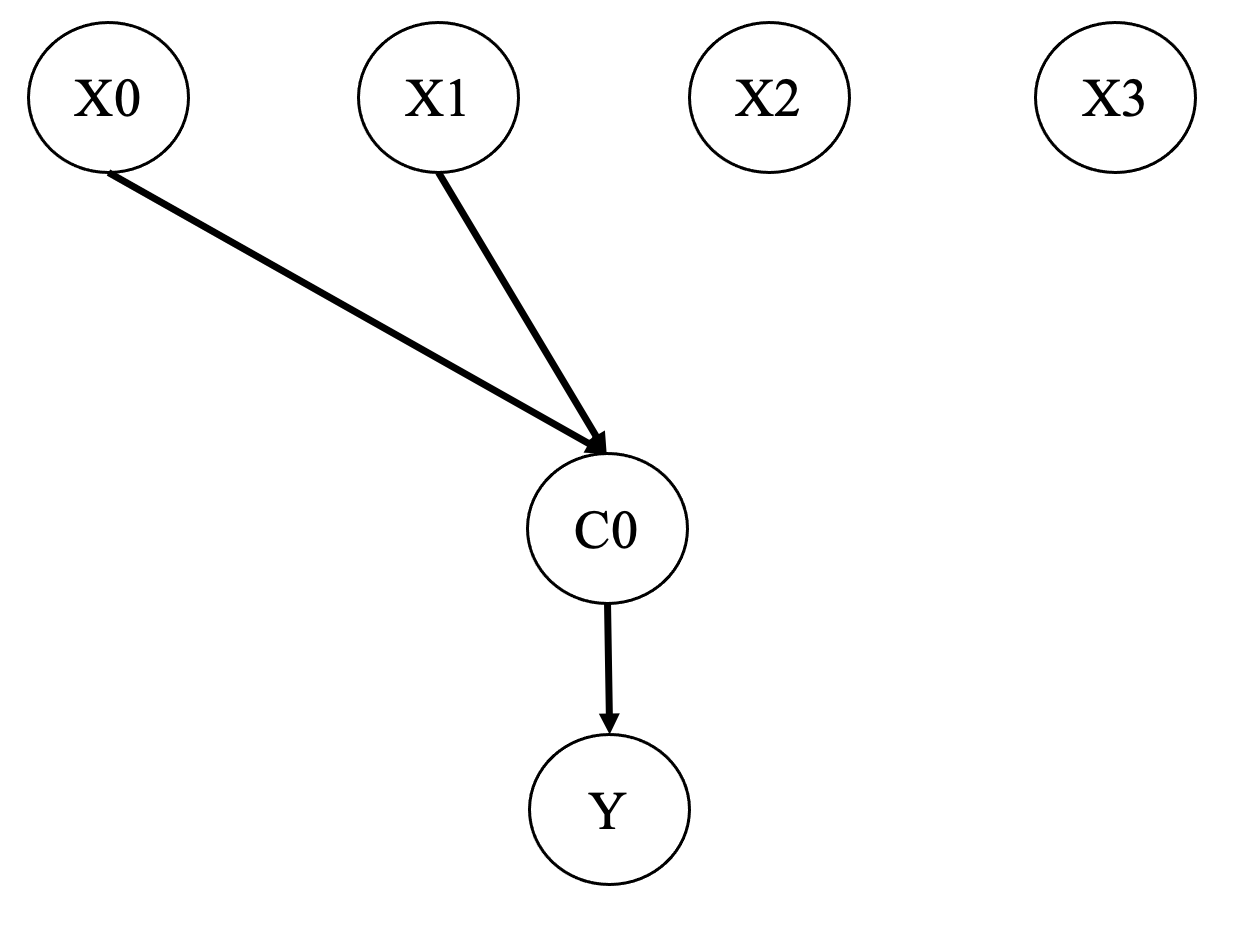}
         \caption{}
         \label{fig:csvform_2}
     \end{subfigure}
     \hfill
     \begin{subfigure}[t]{0.40\textwidth}
         \centering
         \includegraphics[width=\textwidth]{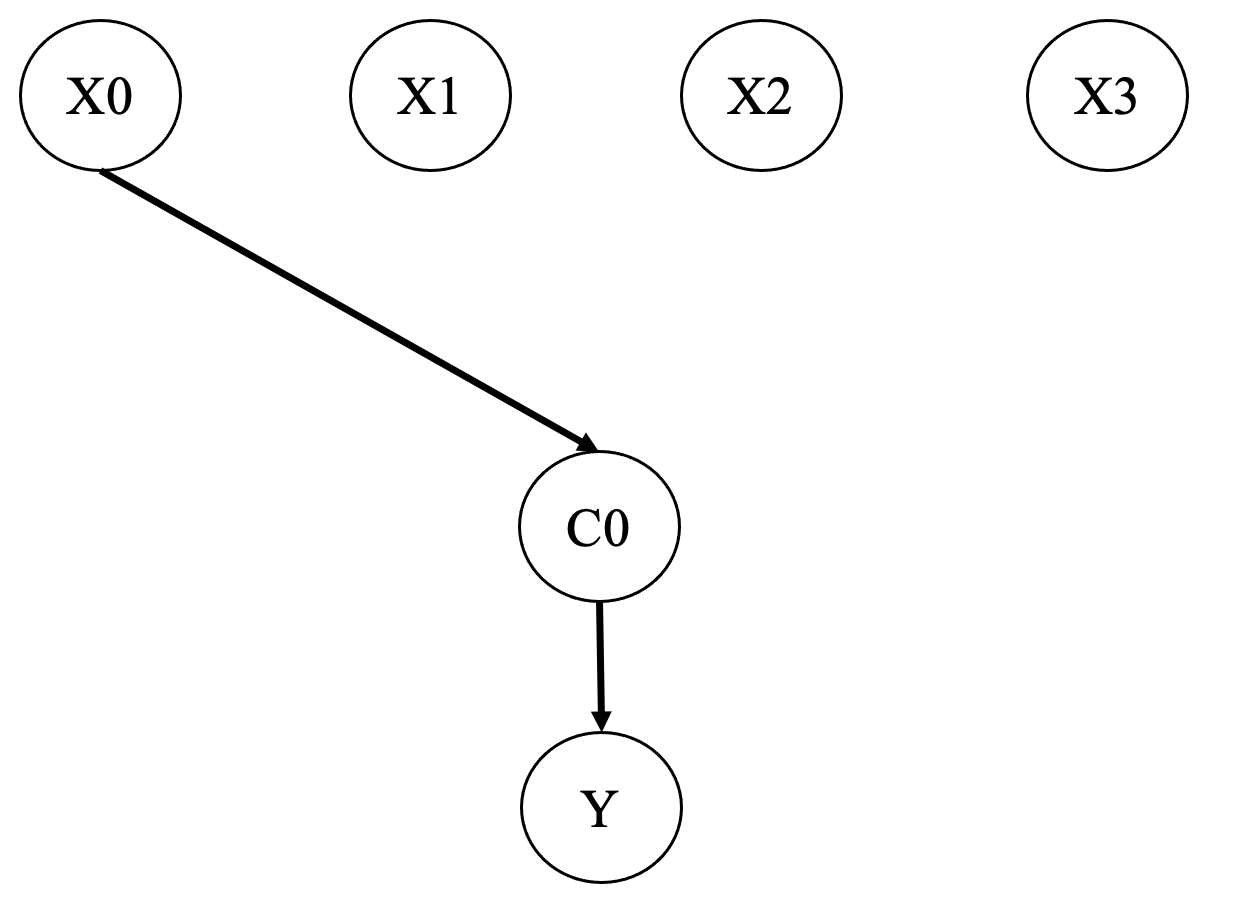}
         \caption{}
         \label{fig:csvform_3}
     \end{subfigure}
     \hfill
     \begin{subfigure}[t]{0.40\textwidth}
         \centering
         \includegraphics[width=\textwidth]{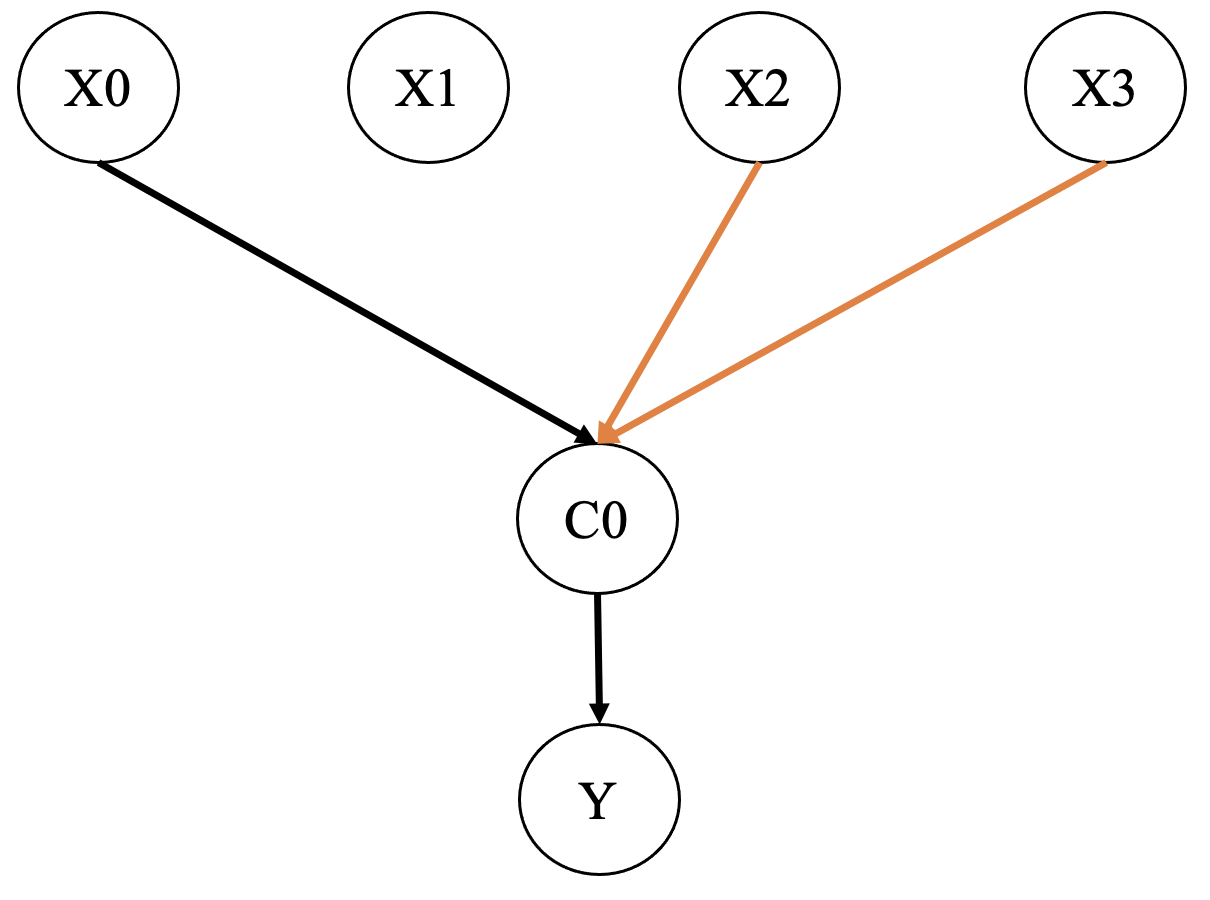}
         \caption{}
         \label{fig:csvform_4}
     \end{subfigure}
     %\hspace{1cm}
     \hfill
     \begin{subfigure}[t]{0.40\textwidth}
         \centering
         \includegraphics[width=\textwidth]{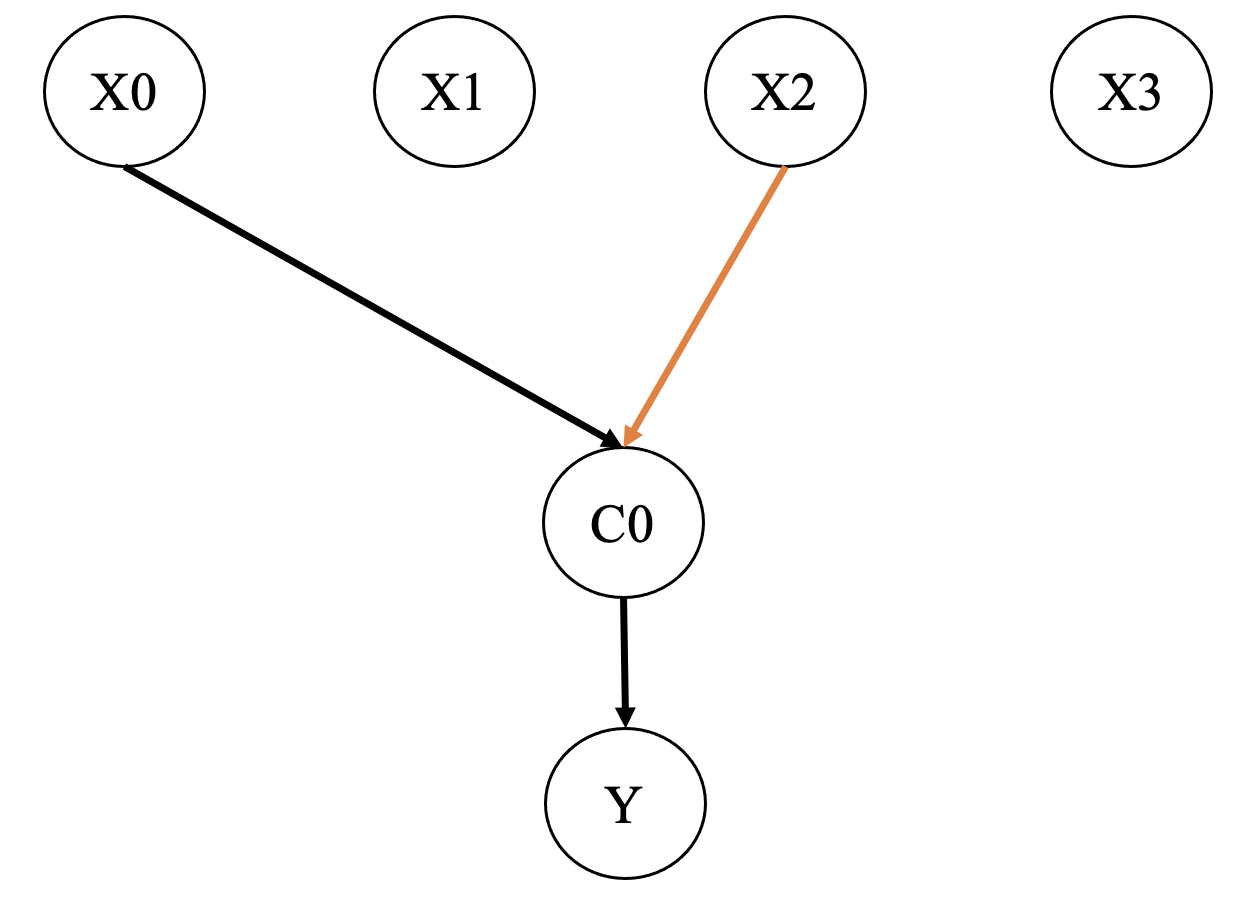}
         \caption{}
         \label{fig:csvform_5}
     \end{subfigure}
     \hfill
     \begin{subfigure}[t]{0.50\textwidth}
         \centering
         \includegraphics[width=\textwidth]{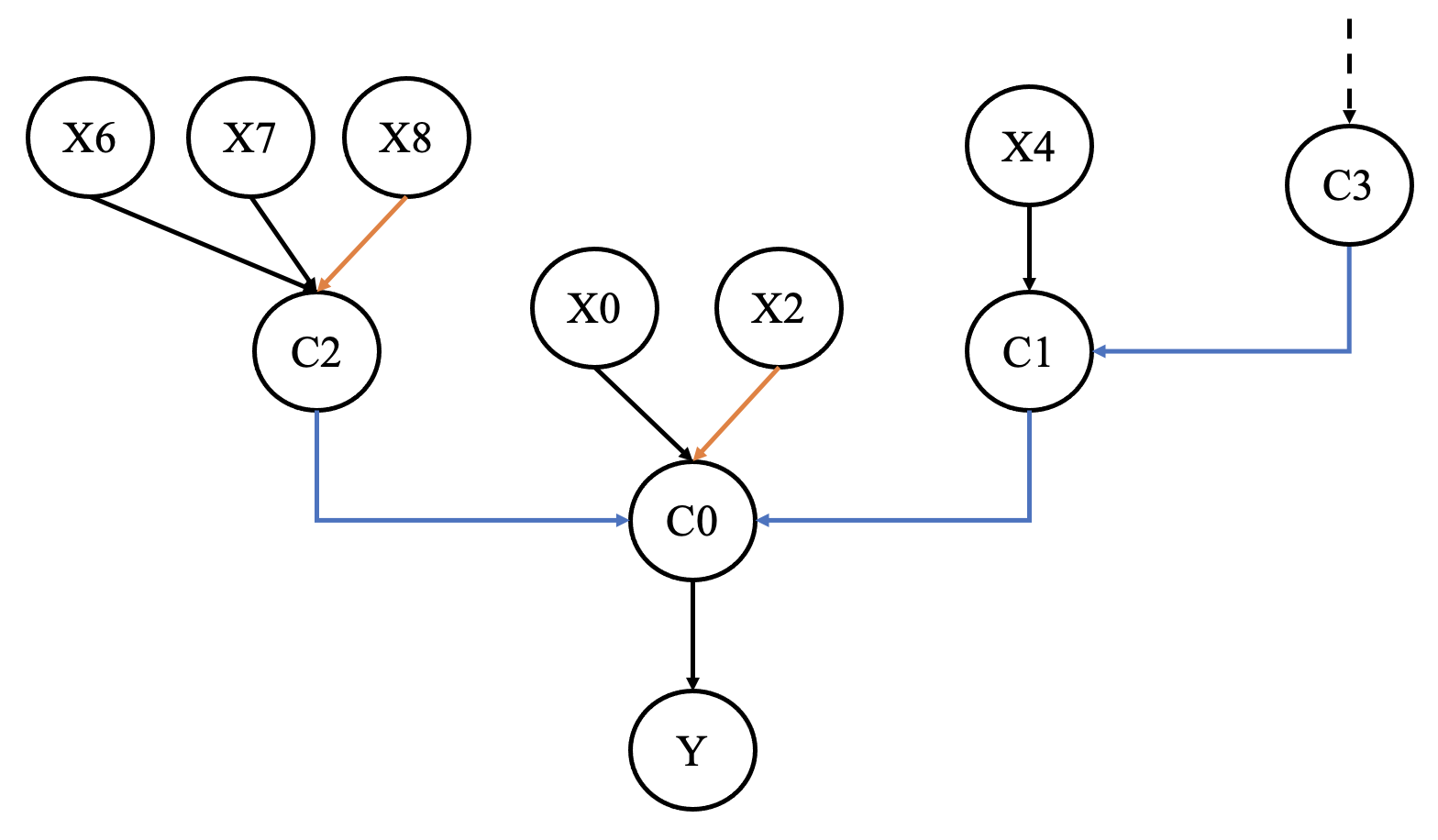}
         \caption{}
         \label{fig:csvformupstream}
     \end{subfigure}
    \caption{Sample formation of a CSV in a continual manner using a component-level variation and selection principle in D2. The relationship to be modelled in (a) to (e) is $Y = X0\ and\ not(X2)$. Black and orange arrows represent positive and negative sources for CSV $C0$ respectively. $Xi$ can be interpreted either as single or grouped SVs. \textit{(a)} Initial state with no relation formed between $X0-3$ and $Y$. \textit{(b)} $X0, X1 \rightarrow Y$ observed. Positive connections hypothesizing both $X0$ \& $X1$ are required for Y are formed. \textit{(c)} $X0 \rightarrow Y$ is observed. $X1$ is deduced unnecessary for $Y$. \textit{(d)} $X0, X2, X3 \rightarrow not(Y)$ observed. $Y$ is hypothesized to be suppressed by $X2$ and $X3$. \textit{(e)} $X0, X2 \rightarrow not(Y)$ observed. $X3$, seen unnecessary for suppression of $Y$, refined. Correct structure learned and is stable from now on. \textit{(f)} A potential upstream conditioning pathway that may follow from (e). The recursive repetition of these processes, where downstream CSVs act as targets for other CSVs, effectively creates points of regulation within the network (akin to D1), enabling the learning of arbitrarily complex relationships.}
    \label{fig:csvform}
\end{figure}

The learning process in this design proceeds step-by-step, without aggregating multiple observational samples from the environment, nor relying on iterative, repeated passes over a data batch—distinct from the majority of contemporary ML approaches. Whenever one or more SVs in the system require an explanation (defined as having an active state without an existing active CSV to explain it), a new CSV is created to account for this. Learning capability of the system comes from the \textit{operations} on CSVs - their formation, and the modification of their positive and negative sources; summarized as follows:

\begin{enumerate}
    \item \textit{Initial formation:} Figure \ref{fig:csvform_2}. At each step, if there are active SVs without an explanation (an active conditioner), a new CSV is generated to explain them. Initially, the CSV has no negative sources ($X_N = \emptyset$) and includes all active input SVs at that timestep as positive sources ($X_P$). No additional positive sources can be added to the CSV.
    \item \textit{Negative connections formation:} Figure \ref{fig:csvform_4}. At the first instance where a CSV's sources are satisfied but its state is inactive, the CSV receives active input SVs at that timestep as negative sources ($X_N$), similar to previous step. No additional negative sources are added thereafter.\footnote{This process is separate from initial sources' formation to avoid creating exhaustive negative connections where unnecessary. Otherwise, a negative connection would be made with everything inactive during CSV creation, which, while accurate, would be overly exhaustive and unnecessary for most negative sources.}
    \item \textit{Refinements:} Figures \ref{fig:csvform_3} and \ref{fig:csvform_5}. When a CSV's state is determined as active with at least one active positive source and active targets, nonactive positive sources (${x \in X_P : S_X \neq 1}$) are removed from $X_P$, and active negative sources (${x \in X_N : S_X = 1}$) are removed from $X_N$.
    \item \textit{Upstream conditioning:} Figure \ref{fig:csvformupstream}. CSVs themselves can serve as targets for other CSVs, enabling recursive upstream repetition of the previously described operations, thereby learning upstream conditioning pathways.
\end{enumerate}

% fig* for single column
% \begin{figure}
%      \centering
%      \begin{subfigure}[t]{0.15\textwidth}
%          \centering
%          \includegraphics[width=\textwidth]{Figures/UpstreamCSV_1.png}
%          \caption{}
%          \label{fig:csvformupstream_1}
%      \end{subfigure}
%      \vline
%      \begin{subfigure}[t]{0.3\textwidth}
%          \centering
%          \includegraphics[width=\textwidth]{Figures/UpstreamCSV_2.png}
%          \caption{}
%          \label{fig:csvformupstream_2}
%      \end{subfigure}
%     \caption{Example of upstream conditioning, continuing from Figure \ref{fig:csvform}. (a) $X0, !X2, X4, X5 \rightarrow Y$ observed. $C0$ is observed to be active, since $XO, !X2$ led to $Y$. A new CSV $C1$ is formed \& conditions $C0$. Note that $(X4, X5)$ alone will not predict activation of $C0$ if $C0$’s sources are not also active. (b) New conditioners are also subject to the CSV processes: Here, the source $X5$ of $C1$ has been refined, and new conditioners $C2$ and $C3$ are formed. Multiple conditioners represent alternative paths: In this case, $C0$ is expected to be active when sources of either $C1$ or $C2$ is active. Any logical function can hence be incorporated in a conditioning pathway in a minimal and ongoing manner without destroying past knowledge.}
%     \label{fig:csvformupstream}
% \end{figure}

Intuitively, a CSV is initially connected to all active SVs at its formation, representing a comprehensive hypothesis of relationships (1, 2) – corresponding to the \textit{local variation} stage. These relationships are then refined based on observations, with unnecessary connections being pruned (3), ensuring the CSV remains locally consistent with past observations and forming \textit{local selection} process in this design. The capability for upstream conditioning (4) enables the representation of arbitrarily complex logical relations in a structurally minimal way. Notably, this latter capability parallels the ENC operation in D1 described above, and as such, CSVs also exemplify potential \textit{points of regulation} — the primary difference from the modulatory nodes in D1 in this regard is that CSVs are present in the system as soon as a new connection is formed, rather than being created on-demand.

The process of initial exhaustive formation, followed by refinement (a component-level variation and selection mechanism) over connections that are defined solely by their presence or absence, rather than fine-tuned weights (weak linkage), underpins the ability for continual learning, in a way that is not imposed from above but is organically present from the lowest levels of organization, as formalized by the following property.

\begin{theorem}
    Let $y_i$ be an \textit{instance} that includes the previous states of all the positive and negative sources of a CSV $C$ and the current states of all its conditioning targets. Then, if $C$ undergoes any modification as a result of encounter with an instance $y_1$, its state in reponse to any past instance $y_0$ is not altered by this modification; as long as its set of targets remain identical and $C$ does not undergo negative sources formation (either because inactive state is not observed or because it has already undergone it).\footnote{The requirement for identicality of targets in this theorem is only to account for the fact that heterogeneous targets result in duplication of CSVs - see the Appendix of \citep{Erden2024Modelleyen} for details of this mechanism. The theorem holds when one considers the response of the duplicated CSVs with respect to the targets assigned to each duplicate as well.}
\end{theorem}

The proof, which can be derived by simply enumerating alternative cases based on source satisfactions, can be found in \citep{Erden2024Modelleyen}. Theorem 2 is exemplified in Figure \ref{fig:csvform}: In \ref{fig:csvform_2}, after elimination of $X1$ as a positive source, the earlier exposure of $X0, X1 \rightarrow Y$ still results in a state of activity in $C0$, and likewise for $X2$ \& $X3$.\footnote{The refined connections can either be removed from the model completely, or, optionally (primarily conditioned on available computational resources), they can be reorganized as the sources of a new upstream CSV that conditions the current one. This forms a hierarchical structure of all observed knowledge in a structurally minimal way, with connections being removed only if the modeled relationships fall below user-defined statistical significance thresholds. If the refined connections are reorganized as upstream conditioners, past \textit{knowledge} is preserved directly. If they are removed entirely, then past \textit{responses} are preserved instead. From a practical/operational perspective, both methods can be considered "continual learning" because the response to a previous observation remains the same in both cases. The former approach simply readapts faster if the refined knowledge proves to be significant for modeling future observations (which, of course, cannot be known in advance without those observations).} With this property, we know that the state of a CSV in response to any past encounter is not altered except possibly during the initial formation of negative sources (which happens only once per CSV). This ensures continual learning without destructive adaptation, inherently and from the lowest level of organization.

\paragraph{Significance}

Design D2 shows that component-level variation-selection (which is the application somatic exploratory processes that form the primary means of adaptation in biological systems to circumstances that cannot be evolutionarily embedded) over weakly-linked components (a key principle in biological systems facilitating reconfigurability) resolves the issue of destructive adaptation observed in statistical learning appraoches in a ground-up manner and by construction. This learning approach is fundamentally different from contemporary methods like neural networks: While contemporary methods initially start with a "bad" solution ("underfitting") and gradually optimize to fit the observed data over time, this design first immediately fits to its observations ("overfitting"), then gradually refines its representations to make them more generalizable, and repeats this process iteratively as needed. Also recall our earlier discussion of neural networks as systems that can only crush an existing pool of variation and not regenerate it (Section \ref{sec:edb_process_varsel}) — in contrast to them, the D2 design repeatedly generates new variation when there is a call for it. While we are not aware of any study or analysis specifically examining somatic exploratory processes based on variation-selection principles from the perspective of long-term preservation of adaptations or learned knowledge (except, perhaps, for time-dependent stabilization mechanisms that straightforwardly achieve this), we believe that their fundamental role in the design presented here offers valuable insights, and may also suggest new directions for analyzing such somatic Darwinian processes. In addition to the benefits of preserving past knowledge, extensions of this design incorporating behavioral mechanisms (\citep{erden2025agential_extendedabs, erden2025agential}) and applications to vision problems (\citep{erden2025mnr}) have demonstrated that such approaches yield intuitive and well-structured learned representations. These representations enhance both direct human comprehensibility and seamless integration with "symbolic" AI methods. Consequently, alongside these extensions, this design, grounded conceptually in EDB, addresses multiple critical challenges in contemporary machine learning simultaneously.

\section{Conclusion}

The prevailing AI design paradigm, despite its widespread success in tackling previously intractable problems, suffers from inherent limitations in key qualitative capabilities essential for effective learning systems. These limitations constrain AI’s potential by preventing the acquisition of new knowledge without erasing past information while also producing unstructured, overparameterized internal representations that hinder human comprehensibility and integration with non-learning processes like deliberation or active information acquisition.

As outlined in this paper, the current state of AI parallels Modern Synthesis, and thus, recent advances in the conceptual understanding of evolution —particularly those driven by evolutionary developmental biology— can provide a unifying foundation rooted in first principles rather than superficial inspirations from biology, enabling a new design philosophy that organically resolves AI’s core limitations. To substantiate these claims, we also demonstrated two learning system designs that exemplify evo-devo principles in action, illustrating the benefits that both AI and evolutionary theory can derive from this perspective.

The connection drawn in this paper, along with its exemplifications here and in the broader AI literature, warrants a call for artificial intelligence to be recognized as an area fundamentally tied to biology in general (not just neuroscience, as has historically been the case; though adopting an evolutionary perspective on AI would likely have ramifications for neuroscience models of an evolutionary or selectionist nature as well  (\citep{edelman-neuraldarwinism-1987, edelman1993neural, fernando2010neuronal, de2015neuronal, fedor2017cognitive, seung2003learning, adams1998hebb, changeux1973theory, loewenstein2010synaptic, calvin1987brain, calvin1998cerebral, fernando2012selectionist})), and specifically to evolutionary theory; as well as a call for greater communication and cross-disciplinary participation between biologists and AI developers/researchers. These two fields might be much more intricately connected than what is currently thought by mainstream research communities.

%\section*{Statements and Declerations}

\section*{Author contributions} Z.D. Erden developed the argumentative \& analytical aspects of this work, designed \& implemented the demonstrations, and wrote the main text of the paper. B. Faltings, as Erden’s doctoral supervisor, provided general guidance, feedback on the progress, and critical input on the manuscript.

\section*{Conflict of interest} The authors declare that they have no conflict of interest.

\section*{Acknowledgements and funding} No external funding was received for this work.




\end{document}